\documentclass[11pt, a4paper, logo, copyright]{googlecloud}

\pdftrailerid{redacted}

\makeatletter
\renewcommand\bibentry[1]{\nocite{#1}{\frenchspacing\@nameuse{BR@r@#1\@extra@b@citeb}}}
\makeatother

\usepackage{kantlipsum, lipsum}
\usepackage{dsfont}

\usepackage[authoryear, sort&compress, round]{natbib}

\usepackage[utf8]{inputenc} %
\usepackage[T1]{fontenc}    %

\usepackage{wrapfig}

\usepackage{multicol}
\usepackage{multirow}
\usepackage[table]{xcolor}
\usepackage{amsmath}
\usepackage{amssymb}
\usepackage{enumitem}
\usepackage{arydshln}
\usepackage[normalem]{ulem}
\usepackage[most]{tcolorbox}
\usepackage{listings}
\usepackage{cuted}
\usepackage{array}

\newcommand{\mypara}[2][\vspace{0.0em}]{#1\noindent\textbf{#2}}

\newcommand{\baseline}[0]{PaperBanana-DirectRefine}
\newcommand{\dataset}[0]{MTPaperBananaBench}
\newcommand{\ours}[0]{PaperBanana-Interact}

\newcommand{\projecturl}{https://shirley-wu.github.io/PaperBanana-Interact/}

\title{\ours{}: Scientific Diagram Refinement with Multi-Turn Human Feedback}

\correspondingauthor{xueqing.wu@cs.ucla.edu, violetpeng@google.com}

\author[1 2 *]{Xueqing Wu}
\author[2]{Ashwin Balasubramanian}
\author[2 3]{Bingxuan Li}
\author[4]{Dawei Zhu}
\author[1]{Kai-Wei Chang}
\author[2]{Yale Song}
\author[2]{Yiwen Song}
\author[2]{Rui Meng}
\author[2]{Tomas Pfister}
\author[2]{Nanyun Peng}

\affil[1]{University of California, Los Angeles}
\affil[2]{Google}
\affil[3]{University of Illinois Urbana-Champaign}
\affil[4]{Peking University}

\begin{abstract}
Recent efforts have aimed to automate scientific diagram generation from paper content \citep{zhu2026paperbanana,lin2026autofigure}. However, fully satisfying an author's visual and communicative preferences in a single turn is challenging: in our formative user study ($N=14$), all participants requested further revisions after viewing an initial draft, and 86\% of them rated the refined diagrams as more satisfactory. Despite the clear demand, the multi-turn workflow remains largely underexplored. To bridge this gap, we present \textbf{\dataset{}}, a benchmark for \textbf{multi-turn diagram generation} containing 292 images annotated with 3,518 user requirements. To reduce expensive human studies and enable scalable benchmarking, we construct a user simulator that, at each turn, identifies unsatisfied requirements and converts $k$ of them into natural language feedback. Evaluating both requirement satisfaction and overall diagram quality reveals two key failure modes shared across baseline multiturn systems: (1) \textit{quality drift}, where diagram quality progressively declines over turns, and (2) \textit{forgetting}, where previously implemented features are lost in subsequent turns. To address these issues, we introduce \textbf{\ours{}}, a multi-agent system that refines diagrams via an internal critique-and-refine loop. \ours{} consistently improves rather than degrades diagram quality across turns, outperforming baselines by 11.9--18.6 points in quality score and reducing forgetting by 3.7--6.2 points.

\end{abstract}

\begin{document}

\maketitle

\section{Introduction}

Visual diagrams are essential for communicating complex scientific ideas. Motivated by this, recent work explores using multimodal generative models to generating diagrams directly from paper content \citep{zhu2026paperbanana,zhu2026autofigure,zhao2026crafter}. However, satisfying an author's nuanced visual and communicative preferences in a single pass remains challenging, necessitating multi-turn refinement. Furthermore, figure design is naturally iterative: reviewing an initial draft often reveals  refinement  opportunities   that are difficult to anticipate upfront. 
Our formative study with 14 participants from diverse backgrounds confirms the  strong  need for a \textbf{multi-turn diagram generation} workflow: all participants requested further revisions after reviewing the initial draft, and 86\% reported higher satisfaction after iterative refinement. These findings highlight multi-turn interaction as a critical yet largely unexplored direction for diagram generation.

To systematically investigate this challenge, we introduce \textbf{\dataset{}}, a benchmark for multi-turn {di}a{g}ram   {r}efinement with iterative user feedback. At the core of the benchmark is the interaction process in which user preferences gradually emerge as the diagram evolves. However, collecting such human interactions at scale is prohibitively expensive and difficult to standardize across evaluation runs. To enable scalable and    reproducible benchmarking, we develop a user simulator that emulates this iterative feedback. The simulator is driven by a set of user preferences hidden from the diagram generator at the initial turn. At each subsequent turn, the simulator identifies unfulfilled preferences and formulates $k$ of them into natural language feedback. To support this simulator, we annotate a diverse pool of 3,518 user requirements derived from 292 diagrams drawn by human experts, spanning content, layout, and visual representation. During evaluation, the generation
\begin{wrapfigure}{r}{0.58\textwidth}
    \centering
    \scalebox{1.03}[1]{\includegraphics[width=0.97\linewidth]{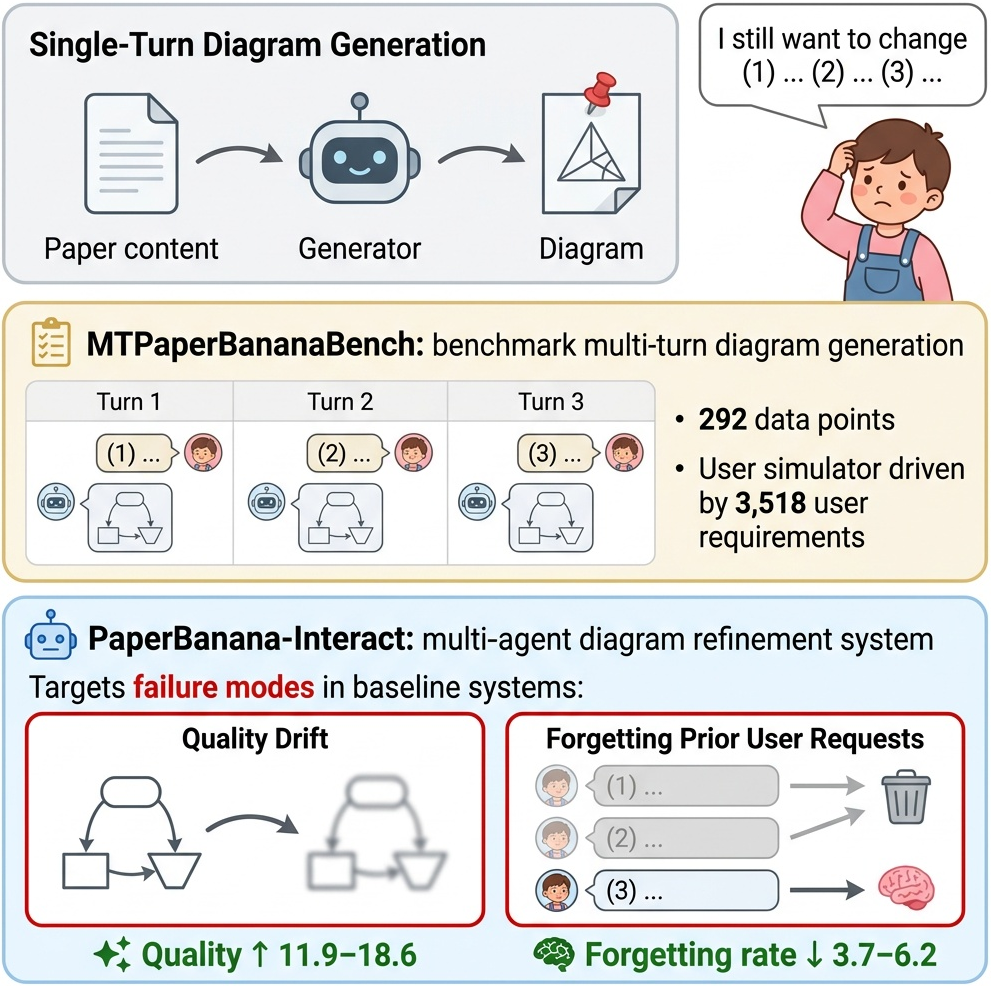}}
    \vspace{-1.3em}
    \caption{Diagram generators may fail to fully satisfy user preferences in a single turn, necessitating multi-turn refinement with human feedback. We introduce \textbf{\dataset{}} to benchmark this task, and propose \textbf{\ours{}} to mitigate common failure modes in baseline systems.~~~~[Figure generated by \ours{}]}
    \label{fig:teaser}
    \vspace{-1.2em}
\end{wrapfigure}
system interacts with the simulated user for a fixed interaction budget. The final output diagram is then evaluated across two dimensions: (1) satisfaction of the user requirements, and (2) overall diagram quality.

Using \dataset{}, we conduct a comprehensive evaluation with various baselines for diagram refinement, including frontier generative models such as NanoBananaPro \citep{google_nano_banana_pro_2025} and GPT-Image-2 \citep{openai2026gptimage2}, as well as an agentic baseline, \baseline{}. While these systems effectively address refinement requests on individual turns, achieving single-turn satisfaction rates of 68.6\% for \baseline{} and 56.2\% for NanoBananaPro, their performance degrades significantly across multi-turn interactions. Extensive failure analysis reveals two key failure modes: (1) \textbf{quality drift}, where image quality progressively deteriorates over turns (50.3$\rightarrow$19.0 for NanoBananaPro, 50.3$\rightarrow$47.1 for \baseline{}); and (2) \textbf{forgetting}, where later refinements overwrite previously satisfied requests, affecting 14.1--22.9\% of such requests across all baselines. These findings highlight the importance of developing diagram generation systems that can effectively interpret evolving user feedback and support iterative refinement across multiple turns.

To address these problems, we propose \textbf{\ours{}}, a multi-agent system for diagram refinement. \ours{} refines the diagram via an internal loop, where a \textit{critic} reviews each revision and proposes suggestions, and a \textit{refiner} and a \textit{visualizer} generate the next revision accordingly. Importantly, \ours{} features a \textbf{multi-objective critic} that assesses the diagram against multiple constraints, including the current user request, all prior user requests,  adherence to the source context, and presentation quality. To handle long, multi-image interaction histories efficiently, we also introduce a \textbf{summarizer} that compresses the full history into a compact \textit{textual memory} capturing relevant visual details, which is then shared across all subsequent agents. Experimental results show that \ours{} mitigates quality drift and forgetting, improving quality scores by 11.9--18.6 percent over baselines while reducing the forgetting rate by 3.7--6.2  percent.

In summary, our contributions are threefold, summarized in Figure \ref{fig:teaser}. (1) We introduce {\dataset{}}, the first multi-turn diagram generation benchmark with 292 samples and 3,518 user requirement annotations. (2) We systematically benchmark generative model baselines and agentic baselines, uncovering shared challenges of {quality drift} and {forgetting} for multiturn scientifc diagram generation. (3) We propose {\ours{}}, a multi-agent refinement system that boosts diagram quality by 11.9--18.6 percent while lowering the forgetting rate by 3.7--6.2 percent.

\section{Related Work}

\mypara[]{Generative models for image generation and refinement.}
Generative  models have achieved strong capabilities in text-guided image generation \citep{cao2025hunyuanimage,google_nano_banana_pro_2025,openai2026gptimage2,meta2026museimage}. %
Beyond one-shot generation, these models increasingly support image-conditioned and instruction-guided refinement, enabling users to modify visual content while preserving desired elements \citep{hertz2022prompt,brooks2023instructpix2pix,kawar2023imagic}. Recent work further explores multi-turn interaction, where users iteratively provide feedback and corrections~\citep{nabati2024personalized,hahn2025proactive}. However, such interaction remains underexplored for scientific diagrams, which impose stricter requirements on structure, semantics, and accuracy. %

\mypara{Scientific diagram generation.} Generating scientific diagrams is challenging because it demands rigorous factual accuracy and the flexibility to capture nuanced author insights. Early methods typically rendered the diagrams by generating code, such as TikZ \citep{belouadi2024detikzify,belouadi2024automatikz,belouadi2025tikzero,zhang2025scimage} or PythonPPTX \citep{zheng2025pptagent,pang2026paper2poster}. With the growing capabilities of generative models, recent research has increasingly integrated them into the pipeline. For example, \citet{zhu2026autofigure} use code to draft an initial diagram before refining it with generative models. Other recent work directly generates the final image with generative models, and relies on large language models (LLMs) solely to optimize the text prompts \citep{zhu2026paperbanana,zhao2026crafter}. Despite these advancements, existing frameworks lack support for multi-turn, iterative human feedback, a critical gap that we aim to address.

\section{Task and Dataset}

\subsection{Task Formulation}
\label{sec:formulation}

In the standard single-turn setting, scientific diagram generation aims to produce an illustrative diagram, denoted as $I_0$, from a source context $x_0$, which typically includes the paper content and a communicative intent that specifies the diagram's scope and focus. The multi-turn setting extends this task by allowing the initial diagram draft $I_0$ to be refined iteratively. At each subsequent turn $t \ge 1$, the user provides natural language feedback $x_t$ requesting further edits. The goal is  then   to generate an updated diagram $I_t$ conditioned on the interaction history $[\mathbf{I}_{<t}, \mathbf{x}_{\le t}]$, where $\mathbf{x}_{\le t}=[x_0,\ldots,x_{t}]$ and $\mathbf{I}_{<t}=[I_0,\ldots,I_{t-1}]$ denote the sequence of user inputs and previously generated diagrams, respectively.

\subsection{Formative User Study}
\label{sec:user_study}

To validate the   practical need for the formulated task, we conducted a formative user study with 14 participants. All participants had substantial experience in academic writing and scientific illustration, with details reported in Appendix \ref{app:user_study}.

\begin{wrapfigure}{r}{0.6\textwidth}
    \centering
    \includegraphics[width=\linewidth]{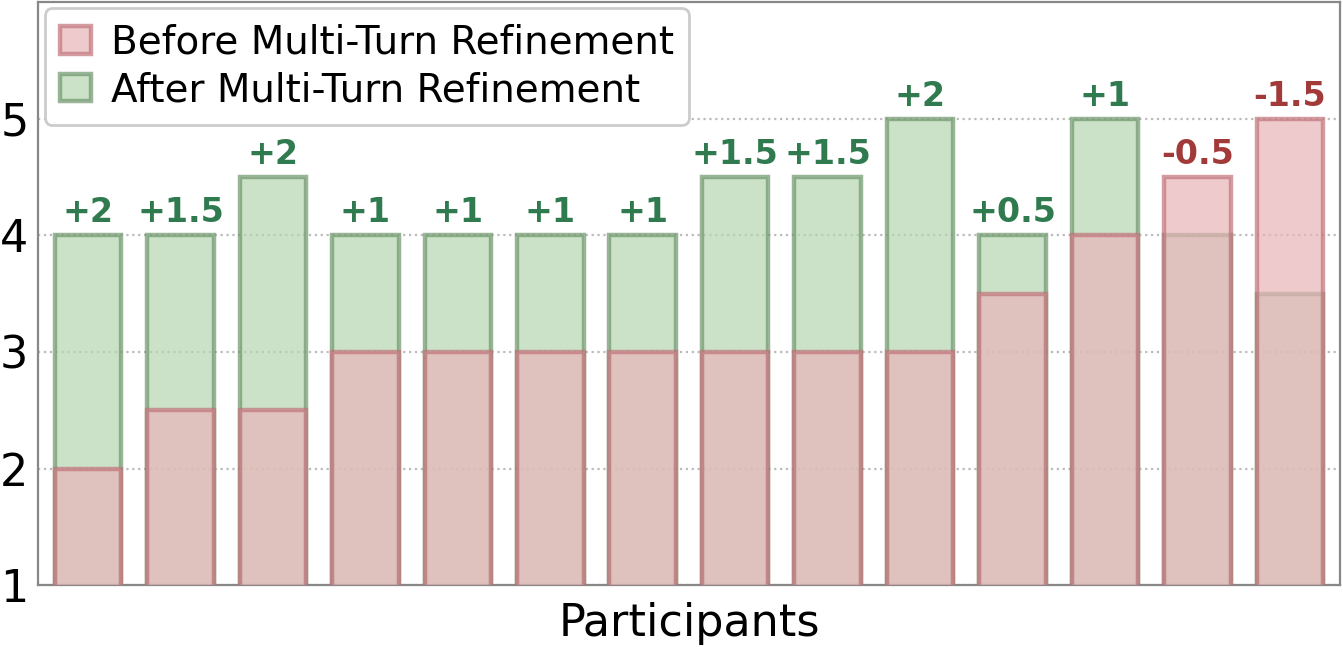}
    \vspace{-1.7em}
    \caption{\textbf{Multi-turn refinement improves user satisfaction.} Per-participant ratings (1--5 scale) for the diagrams \textcolor[HTML]{C1737A}{\textbf{before}} and \textcolor[HTML]{6E966B}{\textbf{after}} multi-turn refinement, with score changes ($\Delta$) annotated above each bar.}
    \vspace{-1.5em}
    \label{fig:user_study}
\end{wrapfigure}

Participants were asked to create a diagram for their own research. They first reviewed an initial draft generated by PaperBanana and then interacted with the baseline system \baseline{} (described in \S\ref{sec:baseline}) as needed to iteratively refine the diagram. Afterward, participants retrospectively rated their satisfaction with both the initial and final diagrams on a 1--5 scale.
As shown in Figure~\ref{fig:user_study}, multi-turn refinement increased the average satisfaction score from 3.2 to 4.2. Overall, 12 of 14 participants (86\%) assigned a higher rating to the diagram after refinement, with improvements ranging from 0.5 to 2 points. These results indicate \textbf{a strong user need for multi-turn interactions} when authoring scientific diagrams.

We further conducted follow-up interviews to analyze the interaction sessions and collect qualitative feedback. Notably, nine participants reported that the initial draft   shaped or changed their 
vision of how the diagram should be presented. Consequently, \textit{many refinement requests emerged only after participants viewed an initial or intermediate draft}, a pattern observed in 64\% of the sessions. This further highlights that multi-turn interaction closely reflects real-world authoring workflows. The interviews also surfaced recurring failure modes of the \baseline{} system, including reduced factual faithfulness (29\%), degraded aesthetic quality (43\%), and failure to preserve previously satisfied requests (29\%). These qualitative observations closely align with our quantitative evaluation on \dataset{}, further motivating the design of \ours{} to address these limitations.

\subsection{\dataset{}}
\label{sec:dataset}

We curate \dataset{} to provide   a controlled benchmark for evaluating multi-turn diagram generation and refinement. To simulate iterative human-AI interaction,  a user simulator operates over a list of pre-annotated requirements $R$, identifying the first $k$ failed requirements  and translating them into natural language feedback. Overall, the benchmark comprises 292 diagrams with 3,518 annotated requirements. We provide   details of the user simulator, the dataset construction process, and our evaluation protocol in this section.

\begin{figure*}[!htbp]
\centering
\scalebox{1.03}[1]{\includegraphics[width=0.95\linewidth]{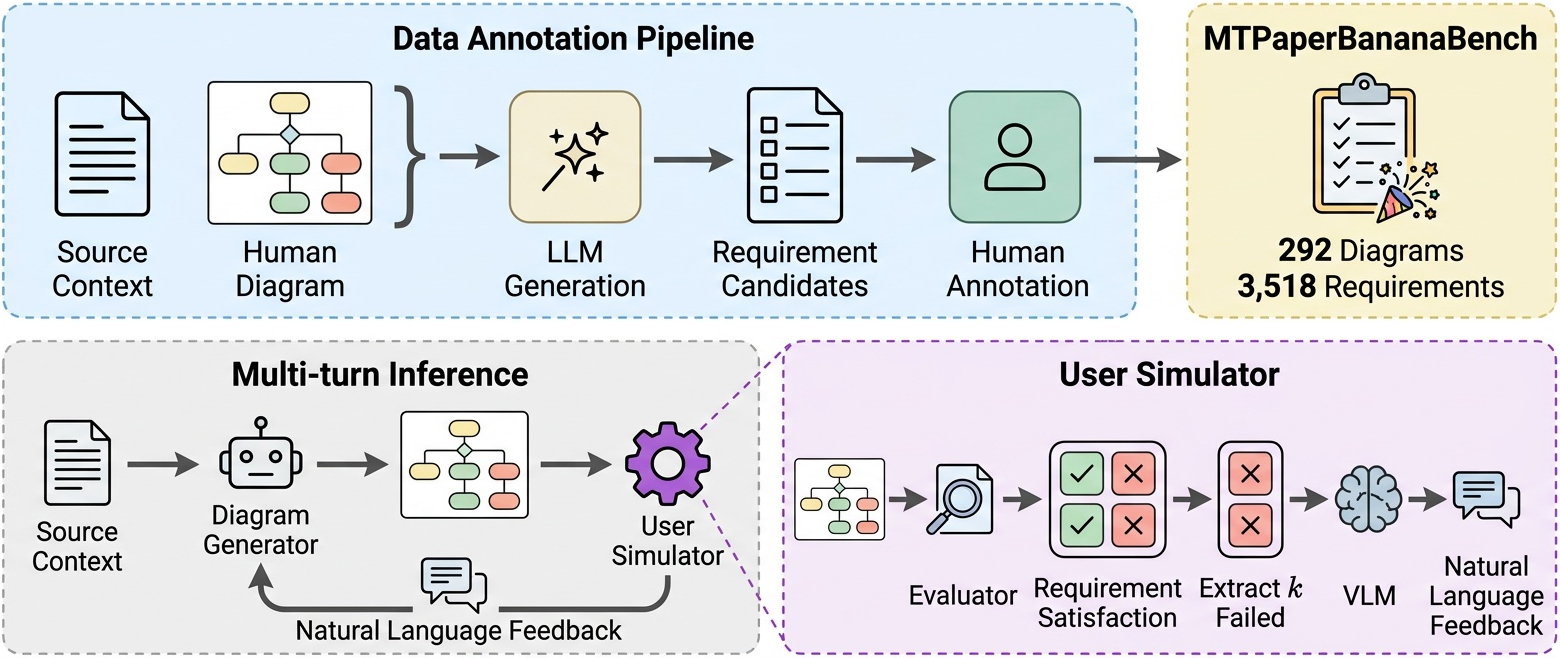}}
\vspace{-.3em}
\caption{\textbf{Annotation and inference pipeline of \dataset{}.} The multi-turn interaction is driven by a user simulator; at each turn, the user simulator evaluates the diagram against all requirements, extracts $k$ failed ones, and formulates them as natural language feedback.~~~~[Figure generated by \ours{}]}
\vspace{-.8em}
\label{fig:user_simulator}
\end{figure*}

\mypara{Dataset construction.}  For each instance in the PaperBananaBench \citep{zhu2026paperbanana} dataset, we annotate a requirement list $R$ through a two-stage process: LLM-assisted candidate generation followed by manual review by human annotators, as illustrated at the top of Figure \ref{fig:user_simulator}. At both stages, the requirement list $R$ is grounded on both its source context $x_0$ and the reference diagram $I_H$ drawn by the original authors. As $I_H$ reflects expert knowledge of the source content and its intended communication goal, we extract its salient  design choices as requirement annotations rather than annotating them from scratch. To ensure broad coverage, we establish a comprehensive taxonomy (detailed in Appendix~\ref{app:annotation}) that categorizes requirements by category (content, organization, and visual representation) and violation severity. Both the LLM and human annotators are then explicitly instructed to annotate requirements spanning a diverse range of categories and severity levels.

The concrete annotation process is as follows, with further details provided in Appendix \ref{app:annotation}. First, we prompt Gemini-3.1-Pro \citep{googledeepmind2026gemini31pro} to generate candidate requirements and label each according to the taxonomy. Then, three human annotators independently review the candidates, removing or revising unfaithful requirements and correcting inaccurate category labels. This manual review removes 0.5\% and revises 24.7\% of the candidates. The final \dataset{} contains 3,518 requirements across 292 diagrams, averaging 12 requirements per diagram, with the detailed distribution shown in Figure~\ref{fig:statistics}. To assess annotation quality, we measure inter-annotator agreement on a subset of 207 requirements, yielding a Cohen's $\kappa$ of 0.767, which indicates substantial agreement. Finally, since the user simulator reveals requirements $R$ in a predefined order, we randomly permute each requirement list and fix the resulting order.

\mypara{User simulator.} The user simulator operates over a list of $N$ requirements, $R=[ r_1, \ldots, r_N ]$, where each requirement represents a plausible visualization preference, such as \textit{``the bar chart must use distinct fill colors to visually distinguish between the two subject types''} or \textit{``the Q, K, and V token sequences must be stacked vertically''}. These requirements are initially hidden from the diagram generation system and are progressively revealed by the simulator, thereby simulating how a user's request evolves over multiple interaction turns.

Concretely, at turn $t$, an LLM-as-judge evaluator $\mathcal{E}$ (detailed at Appendix \ref{app:eval}) determines whether each requirement $r_i \in R$ is satisfied by the latest diagram $I_{t-1}$, yielding a set of failed requirements.
When the set is non-empty, we select the first $k$ failed requirements following the predefined order of $R$, where $k \ge 1$ is a configurable hyperparameter. The user simulator then prompts a vision-language model   to generate an utterance $x_t$ targeting these failed requirements.
To simplify the simulation, we additionally provide the user simulator with the evaluator rationales that detail why $I_{t-1}$ fails each selected requirement. The interaction terminates when $F_{t-1} = \varnothing$ or a turn budget $T_{max}$ is reached.

\mypara{Evaluation.} We evaluate the final diagram   along two main dimensions: \textbf{requirement satisfaction} and \textbf{diagram quality}.
Concretely, we reuse the evaluator $\mathcal{E}$ to compute the requirement satisfaction rate by averaging the binary satisfaction scores across all requirements.
To evaluate diagram quality, we adopt the metric introduced in PaperBananaBench \citep{zhu2026paperbanana}, which compares the generated diagram   against a human-created reference diagram   across multiple  dimensions and reports the resulting win rate. To characterize multi-turn behavior and measure common   failure patterns, we report two additional diagnostic metrics: (1) \textbf{per-turn requirement satisfaction} isolates the system's single-turn instruction-following capability by measuring the fraction of newly raised requirements that are satisfied within their introduction turn; (2) \textbf{forgetting rate} measures the fraction of requirements that are satisfied when introduced but become unsatisfied at any subsequent turn. Further details on all metrics are provided in Appendix~\ref{app:eval}.

\begin{figure*}
    \centering
    \scalebox{1.03}[1]{\includegraphics[width=0.95\linewidth]{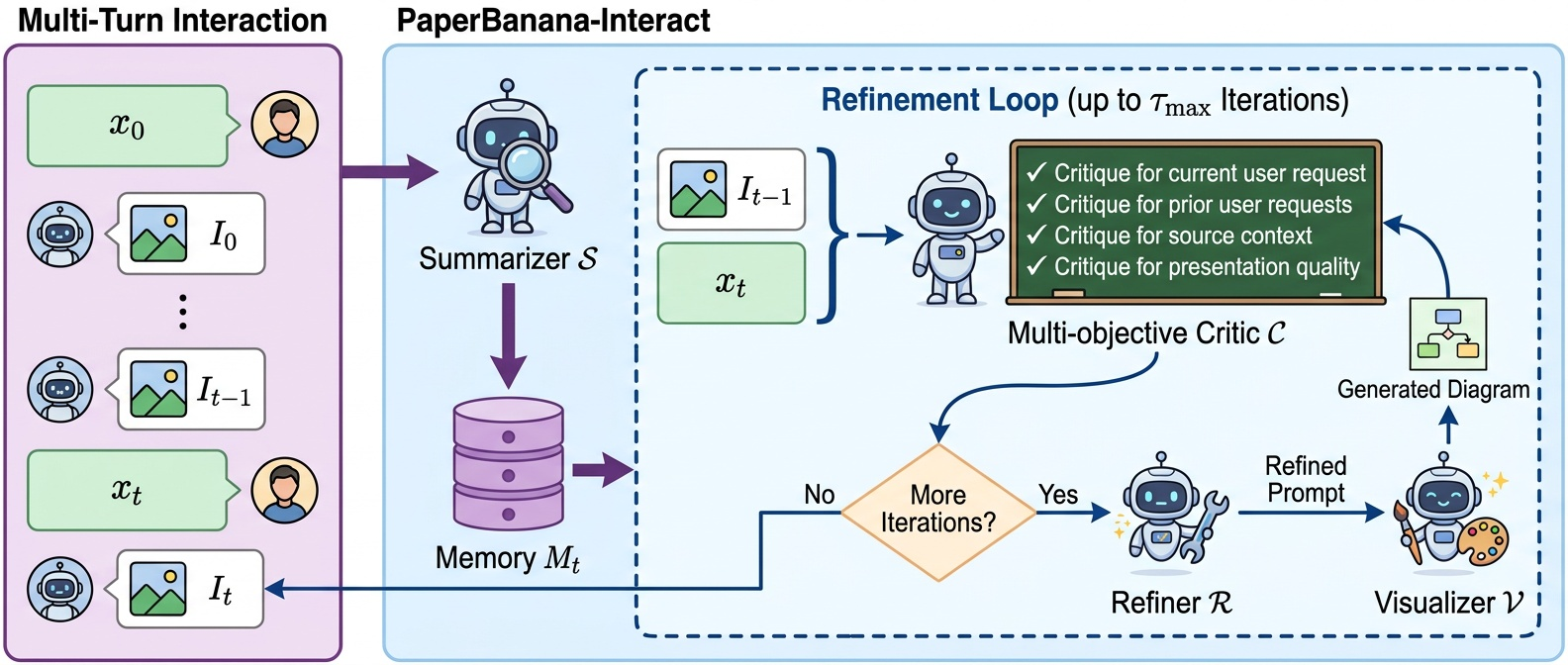}}
\vspace{-.3em}
    \caption{\textbf{\ours{} overview.} \ours{} first employs a \textbf{summarizer} to compress interaction history into a compact memory, followed by a refinement loop where a \textbf{multi-objective critic} evaluates the diagram against different requirements to guide subsequent updates and visualization.~~~~[Figure generated by \ours{}]}
\vspace{-.8em}
    \label{fig:method}
\end{figure*}

\section{Methods}

\subsection{Generative Models as Refiners}
\label{sec:generative_baseline}

Generative models are often inherently capable of image refinement \citep{brooks2023instructpix2pix,kawar2023imagic,openai2026gptimage2,google_nano_banana_pro_2025}, making them a natural baseline for refining diagrams. Denoting the generative model as $\mathcal{V}$, we prompt the model to refine the latest image $I_{t-1}$ conditioned on all user utterances up to turn $t$:
\begin{align*}
I_t = \mathcal{V}({I}_{t-1}, \mathbf{x}_{\le t}).
\end{align*}

\subsection{\baseline{}}
\label{sec:baseline}

Recent agentic diagram-generation systems \citep{zhu2026paperbanana,zhao2026crafter} typically employ a pipeline, denoted by $\mathcal{A}$, to generate and refine a detailed image-generation prompt. A generative model then serves as a visualizer $\mathcal{V}$, producing the final image from the prompt $P$. Under this formulation, the initial turn is given by:
\begin{align*}
P_0 = \mathcal{A}(x_0), ~~ I_0 = \mathcal{V}(P_0).
\end{align*}
Extending this design to a multi-turn setting, the image $I_t$ at turn $t$ is generated by first using a  system $\mathcal{A}_r$ to refine the previous prompt $P_{t-1}$ into $P_t$, and subsequently rendering the image via $\mathcal{V}$:
\begin{align*}
P_t = \mathcal{A}_r(P_{t-1},\mathbf{I}_{<t}, \mathbf{x}_{\le t}),\quad I_t = \mathcal{V}(P_t).
\end{align*}
The effectiveness of this framework thus depends critically on the design of $\mathcal{A}_r$. As a simple baseline, we instantiate $\mathcal{A}_r$ via direct prompting of a vision-language model, denoted by $\mathcal{M}_{\text{vlm}}$. To control computational cost and prevent long-context degradation, we provide the model with the full sequence of user utterances, while limiting the image and prompt inputs to only the most recent turn: %
\begin{align*}
P_t &= \mathcal{M}_{\text{vlm}}(I_{t-1}, P_{t-1}, \mathbf{x}_{\le t}).
\end{align*}

\subsection{\ours{}}
\label{sec:ours}

In this section, we introduce \ours{}, a multi-agent system that builds on the agentic design in \S\ref{sec:baseline}. As shown in Figure~\ref{fig:method}, \ours{} uses a summarizer to compress multi-image histories into a compact textual memory, followed by a critic-refiner loop to iteratively improve both the text prompt and the visualized image. The full prompts are provided in Appendix \ref{app:ours}.

\mypara{Summarizer $\mathcal{S}$.} Requiring each agent to independently process the multi-image interaction history is computationally inefficient and risks distracting it from its core task. Furthermore, successive image revisions exhibit significant visual redundancy and can thus  be substantially compressed. We therefore introduce a summarizer $\mathcal{S}$ that distills the interaction history into a \textbf{textual memory} $M_t = \mathcal{S}(\mathbf{I}_{<t}, \mathbf{x}_{\le t})$.  $M_t$ is shared across all agents, providing them with critical context without requiring repeated processing of the raw images. Concretely, this memory summarizes key visual details, including what each diagram depicts, how it changed relative to the previous revision, and whether those modifications align with the user's request at that turn.

\mypara{Refinement loop.} At turn $t$, \ours{} generates $I_t$  through an internal agent loop starting from the previous turn's image and prompt, i.e. $I_{t}^{(0)} = I_{t-1},~P_{t}^{(0)} = P_{t-1}$. At iteration $\tau$, a  critic $\mathcal{C}$ evaluates the current version $I_{t}^{(\tau-1)}$, identifies remaining defects, and yields a critique:
\begin{align*}
    c_{t}^{(\tau)} = \mathcal{C}\left(I_{t}^{(\tau-1)}, \mathbf{x}_{\le t}, M_t\right).
\end{align*}
If $\mathcal{C}$ determines that further revision is needed, a refiner $\mathcal{R}$ incorporates $c_{t}^{(\tau)}$ to update the prompt:
\begin{align*}
    P_{t}^{(\tau)} = \mathcal{R}\left(I_{t}^{(\tau-1)}, P_{t}^{(\tau-1)}, c_{t}^{(\tau)}, \mathbf{x}_{\le t}, M_t\right).
\end{align*}
The visualizer $\mathcal{V}$ then renders the updated prompt to produce the next image revision, $I_{t}^{(\tau)} = \mathcal{V}\left(P_{t}^{(\tau)}\right)$, which is fed back to $\mathcal{C}$. The loop terminates when the critic determines no further edits are required, or when the  iteration  budget $\tau_{\max}$ is reached.

\mypara{Multi-objective critic.} \ours{} employs a multi-objective critic that systematically evaluates the generated image against all relevant constraints: the current and prior user requests $x_1,\ldots,x_t$, faithfulness to the source context $x_0$, and overall presentation quality. Concretely, the critic is instructed to produce  a structured JSON with an explicit critique entry for each constraint, thereby guiding the system toward satisfying multi-turn constraints and mitigating quality drift and forgetting issues.

\section{Experiments}
\label{sec:experiments}

\mypara[]{Evaluated systems.} To perform  multi-turn diagram generation, we use existing \textit{single-turn generators} to produce initial diagrams and different \textit{refiners} to perform multi-turn iterations. To remain consistent with prior work \citep{zhu2026paperbanana} and ensure a fair comparison, all agentic systems adopt Gemini-3.1-Pro \citep{googledeepmind2026gemini31pro} and NanoBananaPro \citep{google_nano_banana_pro_2025} as the language model and image generation backbones, respectively.

\noindent \textit{Single-turn generators}: We consider generative models, including NanoBananaPro \citep{google_nano_banana_pro_2025} and GPT-Image-2 \citep{openai2026gptimage2}, as well as agentic systems, including AutoFigure \citep{zhu2026autofigure}, AutoFigure-Edit \citep{lin2026autofigure}, PaperBanana \citep{zhu2026paperbanana}, and Crafter \citep{zhao2026crafter}.

\noindent \textit{Refiners}: For generative models used as single-turn generators, we use the same model as the refiner. For agentic single-turn generators such as PaperBanana and Crafter, we compare \ours{} against baseline refiners, including NanoBananaPro and \baseline{}. For \ours{}, we set the iteration budget for its internal loop to $\tau_{max}=10$. We exclude AutoFigure and AutoFigure-Edit from the multi-turn evaluation due to their low performance. %

\newcommand{\grouphead}[1]{%
  \multicolumn{4}{l}{\textit{#1}}\\}

\newcommand{\ablationhead}[1]{%
  \multicolumn{7}{l}{\textit{#1}}\\}

\begin{table}[!htbp]
    \centering
    \begin{tabular}{ll|rrrr}
\hline
\rowcolor{gray!8}Single-Turn Generator & Multi-Turn Refiner & Qual$\uparrow$ & Req$\uparrow$ & Req$^{\text{pt}}\mkern-4.5mu\uparrow$ & Forget$\downarrow$ \\
\hline
AutoFigure & - & 1.9 & 20.4 & - & - \\
AutoFigure-Edit & - & 0.3 & 16.5 & - & - \\
\hline
\multirow{2}{*}{NanoBananaPro} & - & 22.4 & 18.1 & - & - \\
& NanoBananaPro & 12.5 & 39.4 & 53.8 & 20.9 \\
\hline
\multirow{2}{*}{GPT-Image-2} & - & 41.6 & 25.0 & - & - \\
& GPT-Image-2 & 25.7 & 52.9 & 67.4 & 15.5 \\
\hline
\multirow{4}{*}{Crafter} & - & 35.4 & 18.4 & - & - \\
& NanoBananaPro & 6.2 & 39.4 & 56.2 & 19.2 \\
& \baseline{}  & 30.5 & 47.6 & 68.1 & 15.0 \\
& \textbf{\ours{}} & \textbf{48.1} & \textbf{52.1} & \textbf{73.4} & \textbf{10.3} \\
\hline
\multirow{4}{*}{PaperBanana} & - & 50.3 & 25.2 & - & - \\
& NanoBananaPro & 19.0 & 45.2 & 56.2 & 19.0 \\
& \baseline{}~~  & 47.1 & 52.6 & 68.6 & 18.8 \\
& \textbf{\ours{}} & \textbf{61.2} & \textbf{58.0} & \textbf{77.2} & \textbf{12.6} \\
\hline
    \end{tabular}
    \caption{\textbf{Main results with the $k\!=\!1$ user simulator.} We report quality  (Qual$\uparrow$) and requirement satisfaction (Req$\uparrow$) as \uline{main metrics}, and per-turn requirement satisfaction (Req$^{\text{pt}}\mkern-4.5mu\uparrow$) and forgetting rate (Forget$\downarrow$) as \uline{diagnostic metrics}. For each single-turn generator, we report its single-turn performance (- as Multi-Turn Refiner) and its multi-turn performance when paired with different refiners.}
    \label{tab:main}
    \vspace{-.5em}
\end{table}

\mypara{User simulation setup.} We implement the user simulator described in \S\ref{sec:dataset} using Gemini-3.1-Pro \citep{googledeepmind2026gemini31pro} as the backbone model, with full prompts provided in Appendix~\ref{app:user_simulator}. We perform up to $T_{max}=5$ refinement turns with user feedback while allowing early stopping when all user requirements are satisfied. We evaluate two simulation settings with different numbers of requirements raised per turn ($k\!=\!1$ and $k\!=\!3$), reported in Tables~\ref{tab:main} and~\ref{tab:main_3_req}, respectively.

\mypara{Evaluation setup.} As discussed in \S\ref{sec:dataset}, we report two main metrics, requirement satisfaction and diagram quality, both using Gemini-3.1-Pro as the judge. For diagram quality, we adopt the metric from \citet{zhu2026paperbanana} measuring four dimensions: faithfulness, conciseness, readability, and aesthetics. For requirement satisfaction, we prompt the VLM to view the generated diagram alongside its associated requirements and output a binary score and rationale for each requirement. To validate the reliability of this evaluation, a human evaluator independently judged the satisfaction of 200 requirements, achieving a Cohen's $\kappa$ of 0.812 with the VLM judge, indicating near-perfect agreement. For diagnostic purposes, we also report the per-turn requirement satisfaction and forgetting rate. Full prompts and implementation details are in Appendix~\ref{app:eval}.

\subsection{Experimental Results}

\mypara[]{Main results.} We report evaluation results for the $k\!=\!1$ and $k\!=\!3$ user simulators in Tables \ref{tab:main} and \ref{tab:main_3_req}, respectively. As users reveal hidden requirements across turns, multi-turn interaction consistently improves   requirement satisfaction (Req) compared to the single-turn setting. All  refiners are reasonably capable of satisfying a user request in the turn it is raised, with per-turn requirement satisfaction rates (Req$^\text{pt}$) ranging from 50 to 80 across all systems.

\begin{table}[!tbhp]
    \centering
    \begin{tabular}{ll|rrrr}
\hline
\rowcolor{gray!8}Single-Turn Generator & Multi-Turn Refiner & Qual$\uparrow$ & Req$\uparrow$ & Req$^{\text{pt}}\mkern-4.5mu\uparrow$ & Forget$\downarrow$ \\
\hline
\multirow{2}{*}{NanoBananaPro} & - & 22.4 & 18.1 & - & - \\
& NanoBananaPro & 8.6 & 76.5 & 59.6 & 21.5 \\
\hline
\multirow{2}{*}{GPT-Image-2} & - & 41.6 & 25.0 & - & - \\
& GPT-Image-2 & 22.4 & 88.5 & 71.0 & 14.1 \\
\hline
\multirow{4}{*}{Crafter} & - & 35.4 & 18.4 & - & - \\
& NanoBananaPro & 2.9 & 74.6 & 59.1 & 21.6 \\
& \baseline{}~~ & 23.5 & 87.8 & 72.2 & 16.0 \\
& \textbf{\ours{}} & \textbf{42.1} & \textbf{93.4} & \textbf{79.0} & \textbf{12.3} \\
\hline
\multirow{4}{*}{PaperBanana} & - & 50.3 & 25.2 & - & - \\
& NanoBananaPro & 18.3 & 76.9 & 59.8 & 22.9 \\
& \baseline{} & 42.6 & 87.6 & 71.0 & 19.1 \\
& \textbf{\ours{}} & \textbf{54.5} & \textbf{94.2} & \textbf{79.6} & \textbf{13.0} \\
\hline
    \end{tabular}
    \caption{\textbf{Main results with the $k\!=\!3$ user simulator}, with notation following Table \ref{tab:main}.}
    \label{tab:main_3_req}
    \vspace{-.5em}
\end{table}

Beyond this overall trend, performance varies substantially across systems. Among single-turn generators, PaperBanana performs best overall, followed by GPT-Image-2. For multi-turn refiners, generative models such as NanoBananaPro perform poorly and consistently degrade diagram quality. Agentic systems, by contrast, achieve notably better results: \baseline{} significantly improves both diagram quality and requirement satisfaction, while \ours{} delivers further gains.

\mypara{Quality drift.} All baseline refiners exhibit a performance decline relative to their single-turn outputs, 
\begin{wrapfigure}{r}{0.5\textwidth}
    \centering
    \includegraphics[width=\linewidth]{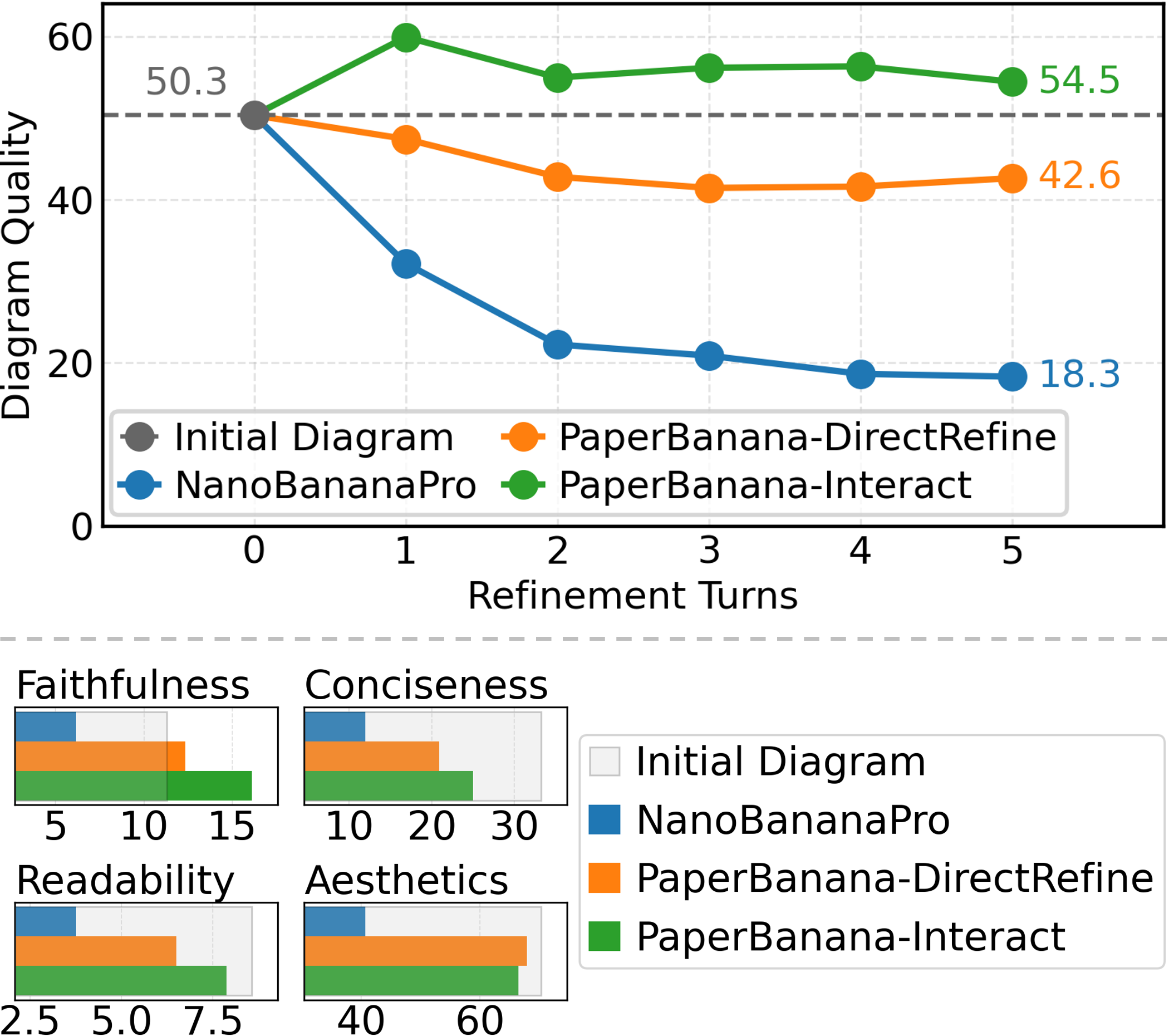}
    \vspace{-1.45em}
    \caption{Diagram quality across refinement turns (top) and final quality breakdown across dimensions (bottom) when refiner systems iterate on PaperBanana outputs for 5 turns with the $k\!=\!3$ user simulator.}
    \label{fig:per_turn_per_dimension}
    \vspace{-1.4em}
\end{wrapfigure}
with generative models experiencing larger drops than \baseline{}. As shown in Figure \ref{fig:per_turn_per_dimension} (top), the quality score experiences a  pronounced drop in early rounds and progressively declines throughout later turns. Comparing Tables \ref{tab:main} and \ref{tab:main_3_req} further shows that this drift is amplified in $k\!=\!3$ settings, where user feedback is longer and contains multiple requirements. In contrast, \ours{} consistently mitigates quality drift and enhances quality over single-turn outputs, despite more modest gains under $k\!=\!3$.

\begin{figure}[!thb]
    \centering
    \includegraphics[width=\linewidth]{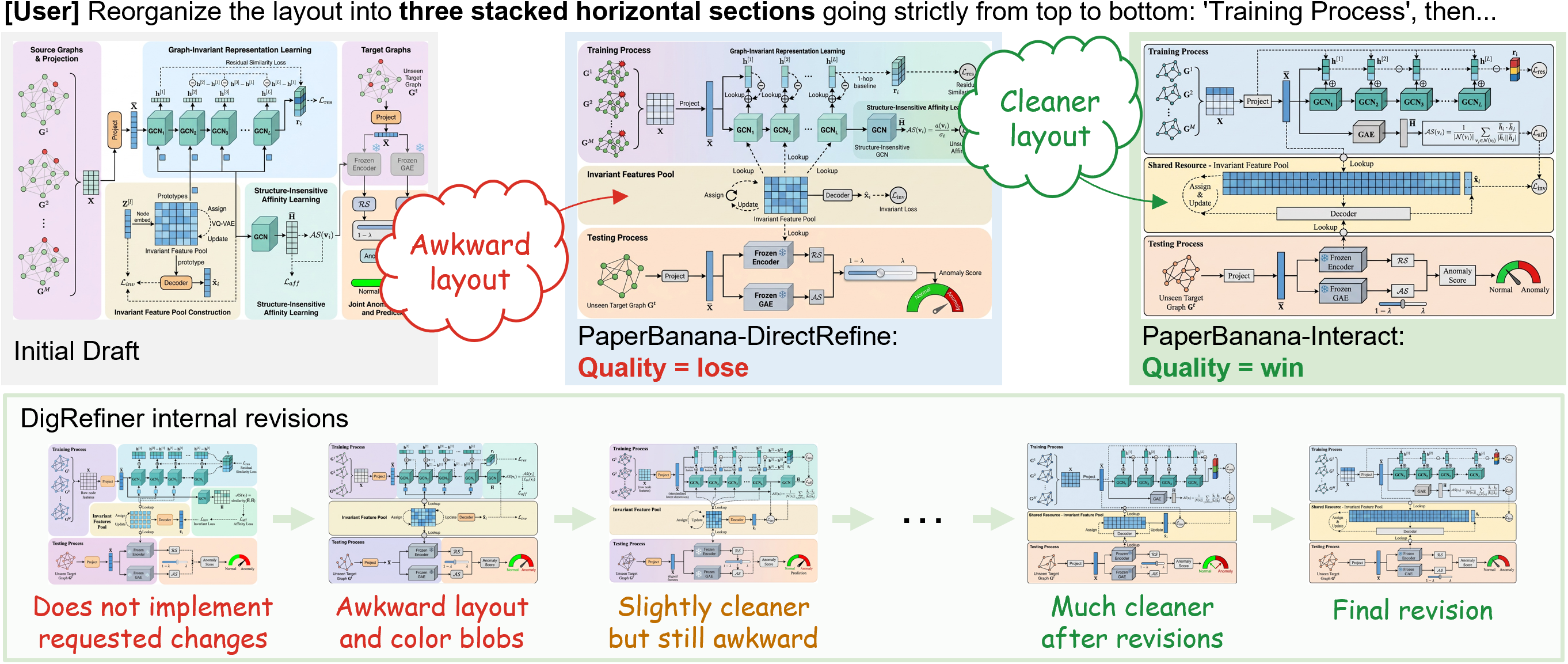}
    \vspace{-1.5em}
    \caption{\textbf{Qualitative example of quality drift.} \baseline{} awkwardly integrates the user request, leaving an unbalanced layout and excessive empty space, while \ours{} iteratively refines the layout and styling and produces a higher-quality output.}
    \label{fig:test_118}
\end{figure}

To dissect quality drift, we present a breakdown of quality scores across dimensions in Figure \ref{fig:per_turn_per_dimension} (bottom). NanoBananaPro exhibits severe quality drift across all dimensions, with prominent failure modes including sudden drops in aesthetics (Figure \ref{fig:test_1}) and distorted artifacts (Figure \ref{fig:test_72}). While \baseline{} better preserves overall quality, its quality drift  often stems from   awkward integration of user requests (Figure \ref{fig:test_118}) and unexpected stylistic drift (Figure \ref{fig:test_112}), mostly degrading conciseness and readability. In contrast, \ours{} mitigates quality drift by iteratively refining the visual design and correcting layout or styling issues across iterations, as shown in Figure \ref{fig:test_118}. A pairwise evaluation by a human annotator across 150 samples further validates these gains: \ours{} outputs were preferred over NanoBananaPro and \baseline{} in 81.3\% and 76.7\% of cases, respectively.

\begin{figure}[!thb]
    \centering
    \includegraphics[width=\linewidth]{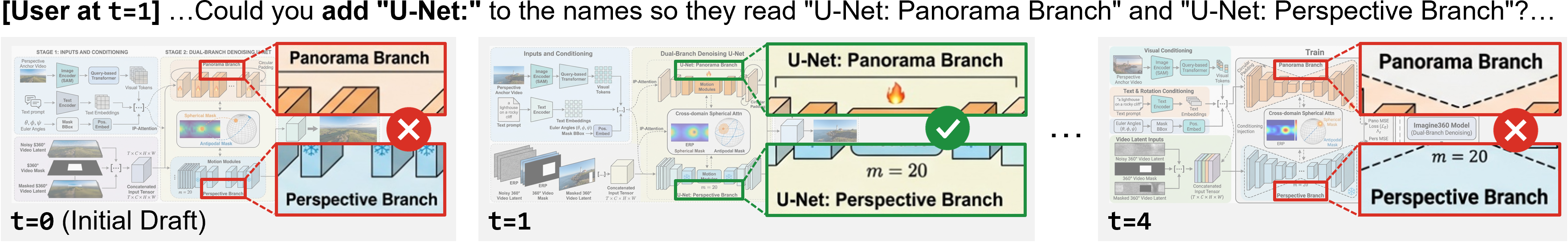}
    \vspace{-1.5em}
    \caption{\textbf{Qualitative example of forgetting.} \baseline{} initially implements the requested changes. However, during later refinements that repeatedly update the styling of the relevant blocks, it fails to preserve the changes.}
    \label{fig:test_120}
    \vspace{-.5em}
\end{figure}

\mypara{Forgetting.}
All baseline systems consistently exhibit forgetting, with forgetting rates ranging from 14 to 23\% under both $k\!=\!1$ and $k\!=\!3$ settings. Forgetting typically occurs when a model optimizes for the current request but inadvertently alters a feature previously specified by the user, as shown in Figure~\ref{fig:test_120}. \ours{} addresses this by refining diagrams over multiple internal iterations while
actively checking for forgetting at each step. As demonstrated in Table~\ref{tab:ablation}, ablated versions with smaller 
\begin{wraptable}{r}{0.57\textwidth}
\small
    \centering
    \begin{tabular}{lll|rr|rr}
\hline
\rowcolor{gray!8}Critic & History & $\tau_{max}$ & Qual$\uparrow$ & Req$\uparrow$ & Req$^{\text{pt}}\mkern-4.5mu\uparrow$ & Forget$\downarrow$ \\
\hline
\ablationhead{Default setup for \ours{}}
MO & Sum & 10 &  \textbf{61.2} & \uline{58.0} & \textbf{77.2} & \uline{12.6} \\[1pt]
\hline
Naive & Sum & 10 & 58.9 & 57.1 & 72.3 & 13.1 \\
\hline
MO & None & 10 & 58.0 & \textbf{58.1} & \uline{77.1} & 14.7 \\
MO & Full & 10 & 58.9 & 57.6 & 73.5 & \textbf{11.6} \\
\hline
MO & Sum & 5 & \uline{60.6} & 57.4 & 75.5 & 13.4 \\
MO & Sum & 3 & 53.3 & 55.9 & 74.1 & 13.8 \\
MO & Sum & 1 & 43.5 & 54.2 & 72.3 & 16.8 \\
\hline
    \end{tabular}
    \caption{\textbf{Ablation study for \ours{}.} We study three design choices: (1) the critic design, comparing our multi-objective critic (MO) with a naive critic that does not enforce structured multi-objective critique (Naive); (2) the multi-image history provided to agents, including the summarized memory $M_t$ (Sum), no history beyond $I_{t-1}$ (None), and the full history $\mathbf{I}_{<t}$ (Full); and (3) the iteration budget $\tau_{max}$ of the internal loop. All runs refine PaperBanana outputs for 5 turns with the $k=1$ user simulator. The highest scores are shown in \textbf{bold}, and the second-highest scores are \uline{underlined}.}
    \label{tab:ablation}
    \vspace{-2.5em}
\end{wraptable}
iteration budgets ($\tau_{\max}$) show monotonically higher forgetting rates. However, \ours{} does not eliminate forgetting entirely, with a forgetting rate of 10--13\%.

\mypara{\ours{} ablations.} We investigate three design choices of \ours{}: critic design, multi-image history, and iteration budget, with results shown in Table \ref{tab:ablation}. For the \textit{critic}, replacing our multi-objective critic with a naive, unstructured baseline lowers per-turn satisfaction by 4.9 points and quality by 2.3 points while slightly increasing forgetting, demonstrating the benefit of our proposed design. For \textit{multi-image history}, omitting it from the input increases forgetting from 12.6 to 14.7 and brings a slight quality drop, confirming that the multi-image history provides critical context for the task. Conversely, feeding agents the full history (\textit{Full}) slightly reduces forgetting but degrades per-turn requirement satisfaction ($77.2 \rightarrow 73.5$) and final quality, while consuming 37.1\% more input tokens. Overall, compact textual memory offers the best trade-off. Finally, the \textit{iteration budget} shows the most significant impact on performance, with a monotonic decline as $\tau_{\max}$ decreases from 10 to 1, where $\tau_{\max}=1$ denotes a single refinement pass  without further  iterations.

\section{Conclusions}

In this work, we investigate multi-turn scientific diagram generation and refinement guided by iterative human feedback. Motivated by a formative user study, we introduce \dataset{} to benchmark this task using a user simulator that progressively reveals requirements across turns. Evaluating requirement satisfaction and diagram quality reveals two key failure modes in existing systems: quality drift across refinement turns and the forgetting of previously satisfied requirements. To address these challenges, we present \ours{}, a multi-agent refinement framework featuring a history summarizer and a multi-objective critic. Empirical results demonstrate that \ours{} consistently mitigates quality drift and reduces requirement forgetting. Overall, this work provides a foundation for future research on interaction-aware scientific diagram generation and refinement.

\bibliographystyle{abbrvnat}
\nobibliography*
\bibliography{custom}

\appendix

\section{Potential Risks}

\begin{figure*}[!tbhp]
    \centering
    \includegraphics[width=\linewidth]{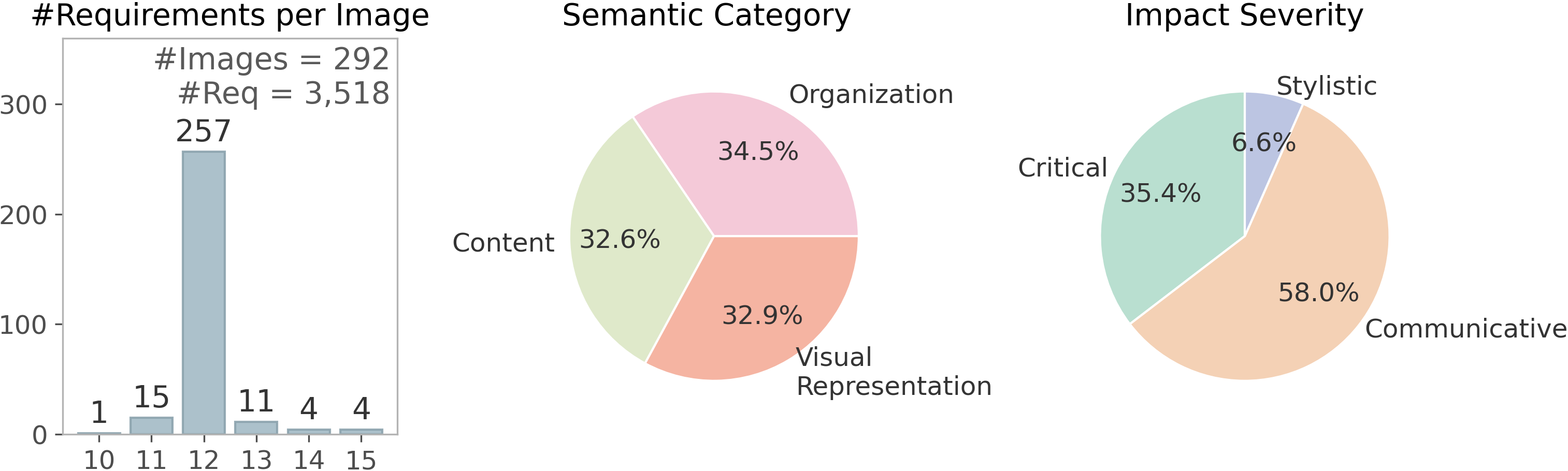}
    \caption{Distribution of the number of requirements per diagram, as well as the distribution of requirements across taxonomy categories  in \dataset{}.}
    \label{fig:statistics}
\end{figure*}

We acknowledge the ethical risks associated with generative models, particularly the potential for ``visual hallucinations'', in which representations of scientific content appear plausible but are factually incorrect. We will explicitly state that the tool is intended solely as an assistive system and will prominently display this disclaimer in both the README file and the main page of the demonstration interface. Following the practices outlined in \citet{zhu2026autofigure} and \citet{lin2026autofigure}, we intend to release the software under a dedicated license requiring that any academic publication using images generated by the tool must  (a) clearly disclose that the image was generated by this system; and (b) include a section discussing the role of artificial intelligence in the research. We emphasize that users should not over-rely on such systems and must maintain rigorous human oversight to preserve the scientific accuracy and integrity of published illustrations.

\section{Data Annotation Details}
\label{app:annotation}

As described in \S\ref{sec:dataset}, we annotate  the requirements using a two-stage process. First, Gemini-3.1-Pro is prompted to generate candidate requirements; human annotators then review and revise them. Both the LLM and the human annotators are provided with the same taxonomy to promote broad and balanced coverage. The taxonomy is defined as follows:
\begin{itemize}[noitemsep,topsep=0pt]
    \item \textbf{Semantic Category:} (1) \textit{Content}, specifying what information appears in the diagram, what is omitted, which functional relationships are shown, and the level of mathematical or functional detail; (2) \textit{Organization}, specifying how information is spatially arranged, ordered, and grouped; and (3) \textit{Visual representation}, specifying the visual encodings, styles, metaphors, and overall aesthetic treatment used to render the diagram.
    \item \textbf{Impact Severity:} (1) \textit{Critical}, where a violation makes the figure scientifically, mathematically, or structurally incorrect; (2) \textit{Communicative}, where a violation preserves technical correctness but makes the figure confusing, overwhelming, or prone to misinterpretation; and (3) \textit{Stylistic}, where a violation affects neither accuracy nor comprehension, but reduces aesthetic quality or professional polish.
\end{itemize}
Annotators are instructed to (1) maintain an approximately even balance across the semantic categories and (2) ensure coverage of all three severity levels.

During manual review, annotators are provided with the following information: (1) the source context from the paper, (2) the human-drawn diagram, and (3) the model-generated requirement description and taxonomy labels. They revise  the requirement annotations to ensure (1) faithfulness to the source context, (2) faithfulness to the diagram, and (3) correct taxonomy labeling. Because our goal is to capture the original authors' design choices, we treat  the human-drawn diagram as the primary source of evidence for visual and structural decisions. To further support the annotation process, we prompt  both Gemini-3.1-Pro and Claude-Opus-4.8 to independently assess the faithfulness of each requirement and verify its taxonomy labels.
Requirements are flagged for closer inspection whenever either model identifies a potential faithfulness issue or the two models disagree on the taxonomy labels. In addition, Gemini-3.1-Pro provides a concise summary of the paper to help annotators understand the relevant technical context.
Finally, since figure captions may unintentionally reveal hidden requirements, we prompt Gemini-3.1-Pro to simplify each caption, removing content redundant with the annotated requirements.

\begin{table*}[!tbhp]
\centering
\small
\begin{tabular}{l|ccc}
\hline
 & \textbf{Min} & \textbf{Max} & \textbf{Median} \\
\hline
Papers authored or significantly contributed to in total
    & 3 & 30 & 11 \\
Papers authored or significantly contributed to in the past two years
    & 1 & 25 & 8 \\
    \hline
Paper diagrams personally created or substantially edited in total
    & 2 & 50 & 10.5 \\
Paper diagrams personally created or substantially edited in the past two years
    & 1 & 25 & 6.5 \\
\hline
\end{tabular}
\caption{Summary statistics for participants' prior experience in paper drafting and diagram drawing.}
\label{tab:user_study_statistics}
\end{table*}

In total, three annotators, all authors of this paper, participated in the annotation process. We began with annotation training, followed by the joint annotation of 15 data points, during which discrepancies were discussed and resolved immediately. The annotators then worked independently on separate batches containing overlapping samples, enabling us to compute inter-annotator agreement (IAA). As reported in \S\ref{sec:dataset}, we computed IAA on a challenging subset of 207 requirements that were flagged as potentially problematic by at least one of Gemini-3.1-Pro or Claude-Opus-4.8. On this subset, the three annotators achieved substantial agreement, with a Cohen's $\kappa$ of 0.767, supporting the reliability of the annotations. Detailed statistics of the final dataset are reported in Figure \ref{fig:statistics}.

\section{Details for Formative User Study}
\label{app:user_study}

In this section, we report the detailed protocol for our formative user study. We recruited 14 participants internally from the same organization, with national backgrounds including the United States, China, and India. All participants had substantial experience in academic writing and scientific illustration. Specifically, participants were required to have (1) authored or made significant contributions to at least one research paper, and (2) personally created or substantially edited at least one paper diagram within the past two years. Statistics of participants' self-reported experience are shown in Table~\ref{tab:user_study_statistics}. We obtained consent from all participants to use the results to motivate our study design and to report de-identified results in the final paper.

During the study, each participant selected a paper for the diagram creation task. To ensure sufficient familiarity with the content, the selected paper was required to be one to which they had made substantial intellectual contributions. We encouraged participants to use an \textit{ongoing paper} whose methodology section was largely complete but whose main figure had not yet been created. If no such paper was available, participants instead selected a \textit{previously published paper} and were instructed to adopt a \textit{post-hoc mindset}: rather than reproducing the original figure, they were asked to imagine redesigning or revising it for a keynote presentation or journal extension, with the goal of communicating the core idea more effectively. Among the 14 participants, 5 selected ongoing papers and 9 selected prior papers.

\begin{figure*}[!tbh]
        \centering
        \includegraphics[width=\linewidth]{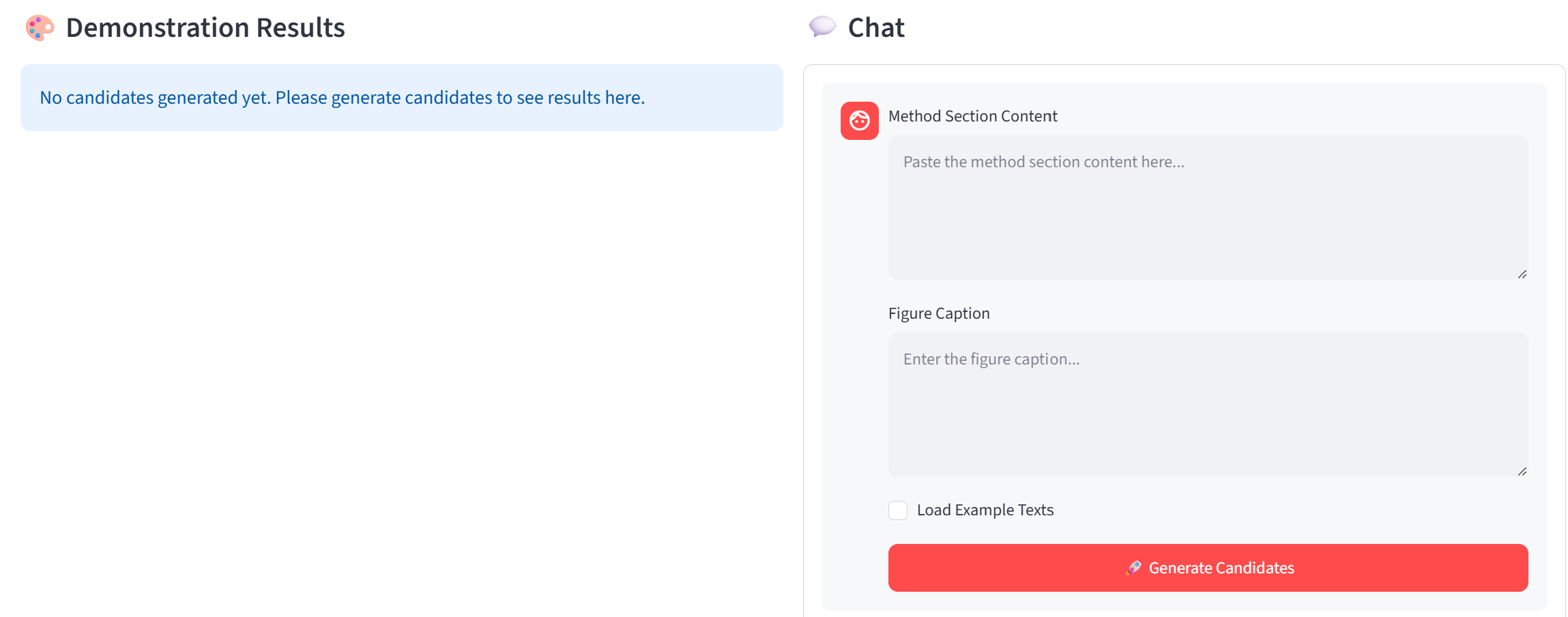}
        \caption{Interface for the initial diagram generation in the formative user study. Participants provided the methodology section of their selected paper and a figure caption describing the intended diagram.}
        \label{fig:user_study_interface_1}
\end{figure*}

\begin{figure*}[!tbh]
        \centering
        \includegraphics[width=\linewidth]{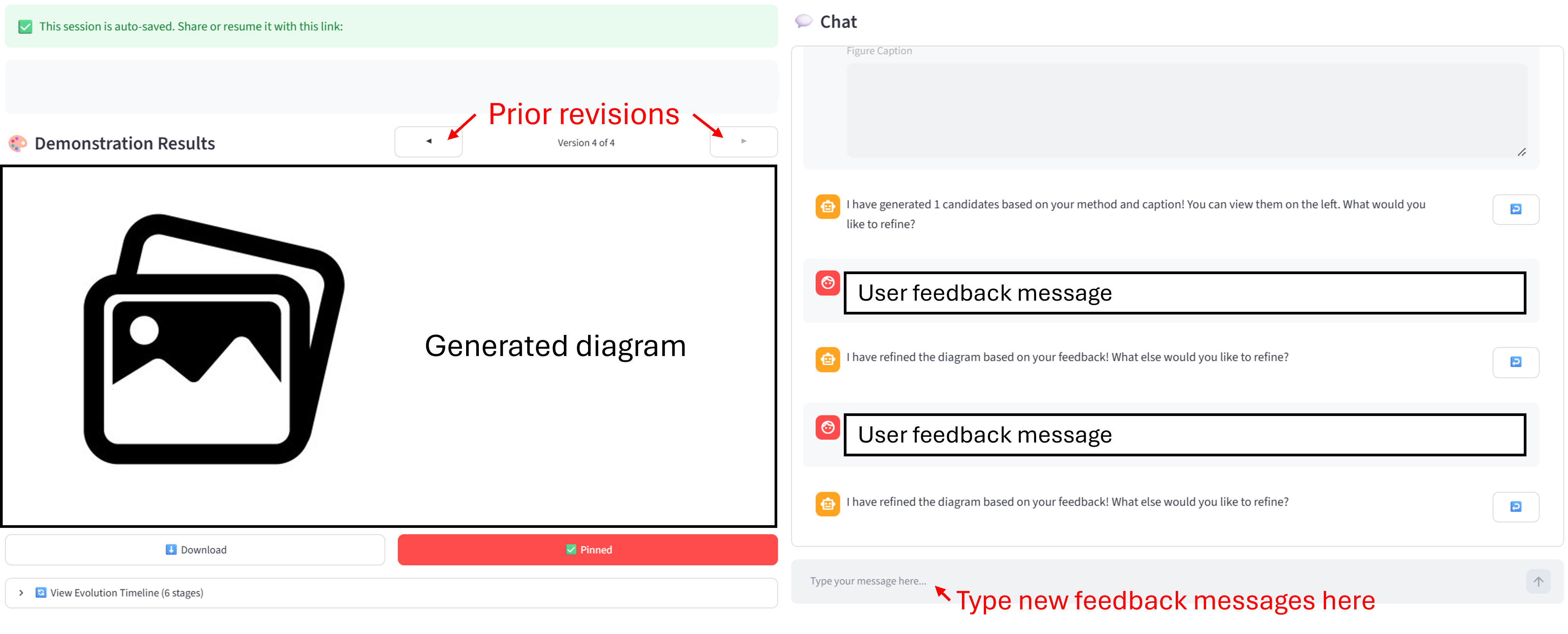}
        \caption{Interface for  multi-turn diagram refinement in the formative user study. The generated diagram and its prior revisions are displayed on the left, while the chat panel on the right allows participants to iteratively refine the diagram through natural-language feedback. User-related information is hidden.}
\label{fig:user_study_interface_2}
\end{figure*}

\begin{figure*}[htbp]
    \centering

    \begin{minipage}{0.66\textwidth}
        \centering
        \includegraphics[width=\linewidth]{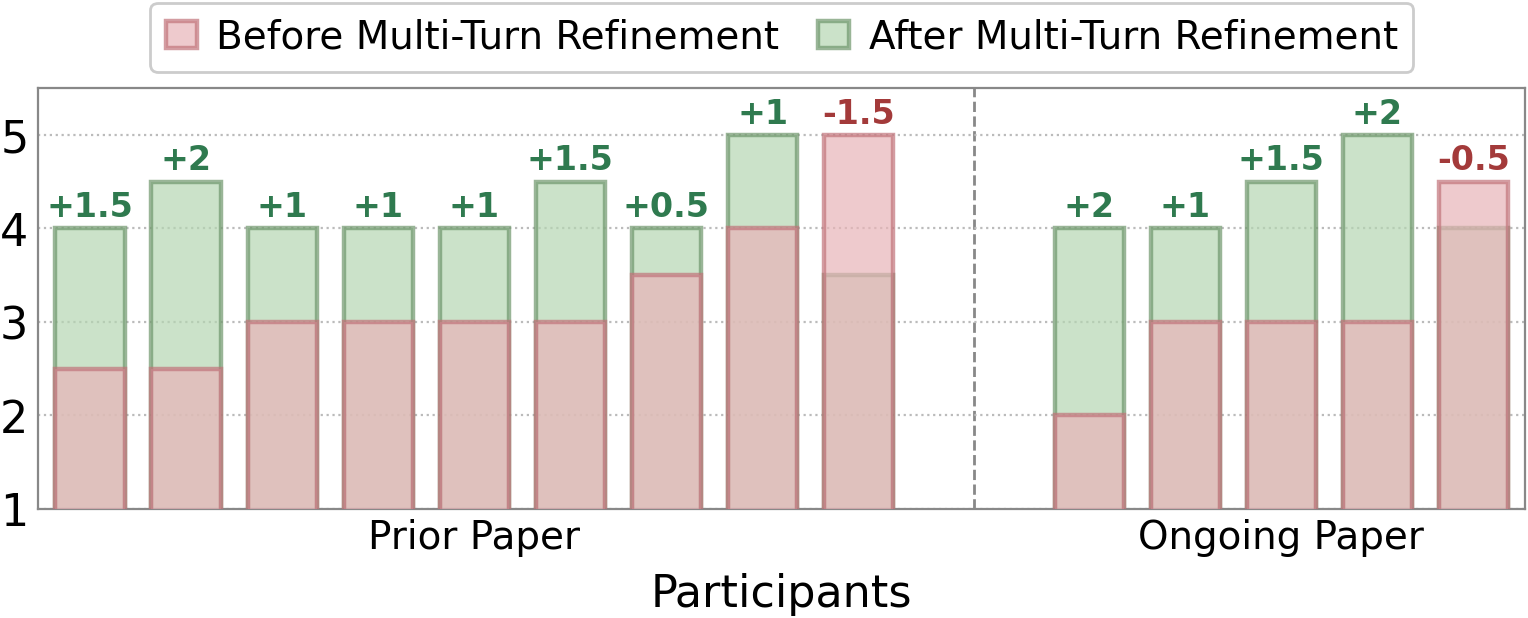}    \caption{Per-participant ratings (1--5 scale) for the diagrams \textcolor[HTML]{C1737A}{\textbf{before}} and \textcolor[HTML]{6E966B}{\textbf{after}} multi-turn refinement, split according to whether participants designed diagrams for prior or ongoing papers. Score changes ($\Delta$) are annotated above each bar.}
    \label{fig:user_study_by_ongoing_or_prior}
    \end{minipage}
    \hfill
    \begin{minipage}{0.31\textwidth}
        \centering
        \includegraphics[width=\linewidth]{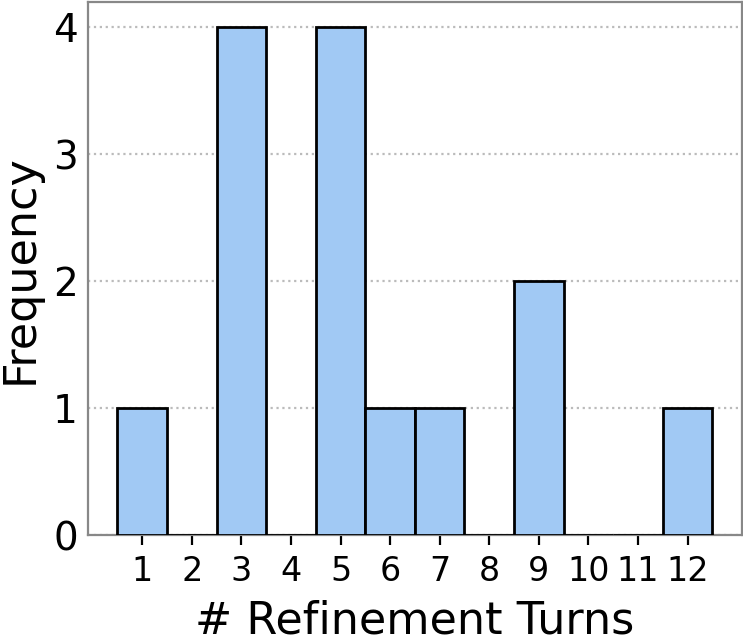}   
        \caption{Number of refinement turns per participant  before they stopped interacting with the system.}
    \label{fig:user_study_stopping_round}
    \end{minipage}
\end{figure*}

After paper selection, participants provided the methodology section together with a figure caption describing the intended purpose of the diagram. The interface for this stage is shown in Figure~\ref{fig:user_study_interface_1}. PaperBanana then generated an initial diagram. Participants could subsequently interact with \baseline{} to iteratively refine the result using natural language feedback. They were free to stop whenever they felt satisfied, with a maximum study duration of 30 minutes. The interaction interface is shown in Figure~\ref{fig:user_study_interface_2}. Every participant requested at least one revision. Participants stopped after a median of 5 interaction turns, with the full distribution shown in Figure~\ref{fig:user_study_stopping_round}.

Following the interaction session, we conducted a semi-structured interview consisting of the following questions:
(1) Rate your satisfaction with the initial diagram on a 1--5 scale.
(2) Rate your satisfaction with the final diagram on a 1--5 scale.
(3) Did seeing the initial diagram change or shape how you thought the method should be presented?
(4) For each revision turn, why did you make this request?
(5) For each revision turn, did the resulting diagram introduce any new problems? If so, why?

Overall satisfaction scores are summarized in Figure \ref{fig:user_study}. Figure \ref{fig:user_study_by_ongoing_or_prior} further separates the results by whether participants used ongoing or prior papers; we observe no significant difference in the distributions between the two groups. Overall, participants consistently reported higher satisfaction with the refined diagrams than with the initial generations.

The interview responses  provide further insight into the iterative design process. Nine of the 14 participants (64\%) reported that viewing the initial diagram changed or refined their thinking about how the method should be presented. This observation is further reflected in participants' explanations of their revision requests. When asked ``Why did you raise this request?'', participants in 9 of the 14 sessions (64\%) reported at least one revision that was prompted by new ideas about how or what to illustrate after viewing the previous diagram, rather than merely correcting an existing deficiency.

The interviews also reveal a complementary phenomenon: the iterative refinement process itself could introduce new problems. When asked whether a revision result had introduced another issue, participants in 11 of the 14 sessions (79\%) reported at least one such occurrence. The detailed breakdown of these issues is reported in the main paper: reduced factual faithfulness in 29\% of sessions, degraded aesthetic quality in 43\% of sessions, and failure to preserve previously satisfied requests in 29\% of sessions.

\section{Experimental Details}

For Gemini-3.1-Pro, we use the \texttt{gemini-3.1-pro-preview} model with a temperature of 1.0 and a permissive maximum output length of 50,000 tokens. For NanoBananaPro, we use the \texttt{gemini-3-pro-image} model with a temperature of 1.0. Following the respective settings used in the official PaperBanana \citep{zhu2026paperbanana} and Crafter \citep{zhao2026crafter} implementations, we generate Crafter images with a 16:9 aspect ratio at 2K resolution. For PaperBanana, we use the aspect ratio specified by each data point and generate images at 1K resolution. For GPT-Image-2, we use the \texttt{gpt-image-2} model with the image size set to 1536 × 1024, quality set to high, background set to opaque, and output format set to PNG.

\begin{wrapfigure}{r}{0.5\textwidth}
    \centering
    \includegraphics[width=\linewidth]{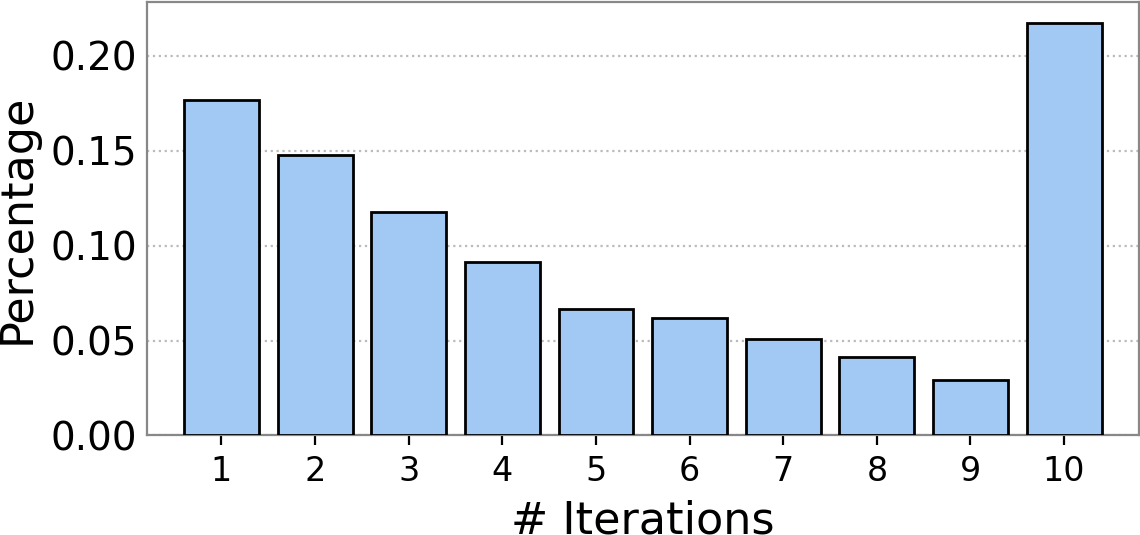}
    \vspace{-1.5em}
    \caption{Distribution of the number of internal agentic iterations performed by \ours{}.}
    \label{fig:ours_iter_stop}
    \vspace{-1em}
\end{wrapfigure}
We then report the computational cost of generating an initial diagram with PaperBanana and refining it with \ours{} ($\tau_{\max}=10$) for five turns using the $k\!=\!1$ user simulator. For each diagram, the initial generation round consumes 186.8$k$/6.2$k$ input/output tokens for Gemini-3.1-Pro and 5.7$k$/5.3$k$ input/output tokens for NanoBananaPro. In each subsequent refinement turn, \ours{} consumes 159.4$k$/11.2$k$ input/output tokens for Gemini-3.1-Pro and 7.1$k$/5.0$k$ input/output tokens for NanoBananaPro per diagram. Most of the computational overhead arises from the internal refinement iterations. Figure~\ref{fig:ours_iter_stop} shows the distribution of the number of iterations performed by \ours{} before stopping. The distribution is concentrated at small iteration counts and generally decreases as the number of iterations increases, with a  spike (22\%) at the maximum iteration limit, corresponding to challenging examples  that cannot satisfy the stopping criterion within the given budget.

Due to computational constraints, we do not report full descriptive statistics for every experiment; most numbers in the main text are therefore from a single run. To estimate the variance, we rerun one of our main settings, refining PaperBanana outputs (fixed throughout five runs) with \ours{} using the $k\!=\!1$ user simulator. Across  five runs, we obtain (mean $\pm$ SD) a diagram quality of 61.2 $\pm$ 1.8, requirement satisfaction of 58.0 $\pm$ 0.4, per-turn requirement satisfaction of 77.2 $\pm$ 0.3, and forgetting rate of 12.6 $\pm$ 1.2. Diagram quality is slightly less stable, likely due to the subjective nature of the quality definition, while the other metrics show low variance. We report the mean values in Tables~\ref{tab:main} and~\ref{tab:ablation} in the main text.

\section{Qualitative Examples}

\begin{figure*}[!pthb]
    \centering
    \includegraphics[width=\linewidth]{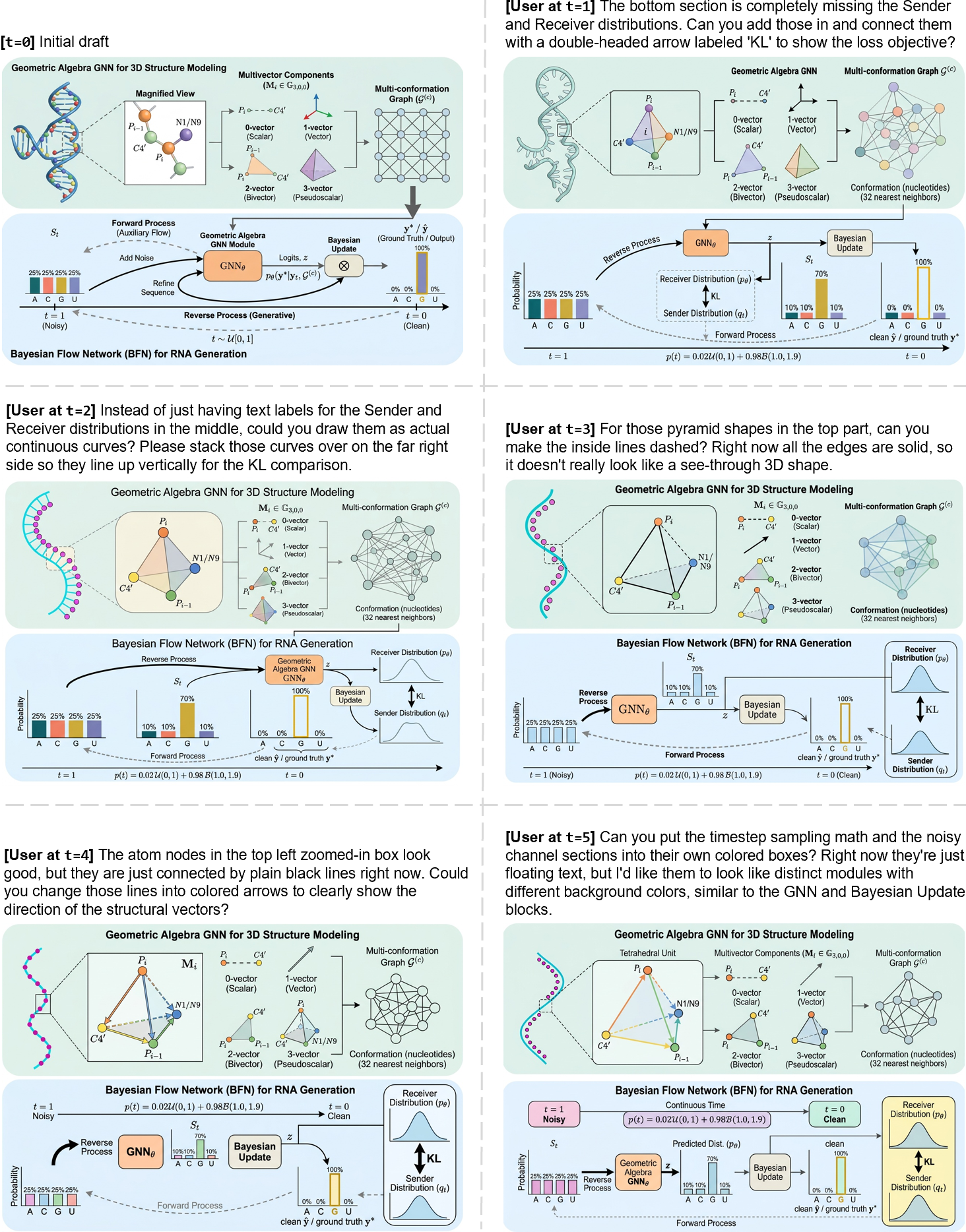}
    \caption{A five-turn refinement trajectory of \ours{} starting from the PaperBanana output. User feedback is produced by the $k\!=\!1$ simulator.}
    \label{fig:full_example}
\end{figure*}

\begin{figure*}[!tbh]
    \centering
    \includegraphics[width=\linewidth]{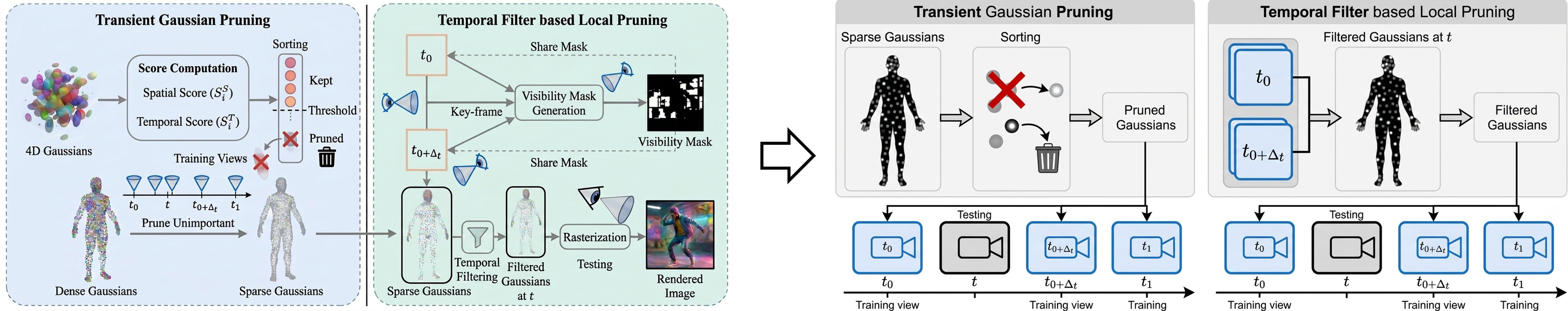}
    \caption{NanoBananaPro refiner outputs with sudden drops in aesthetics.}
    \label{fig:test_1}
\end{figure*}

\begin{figure*}[!tbh]
    \centering
    \includegraphics[width=\linewidth]{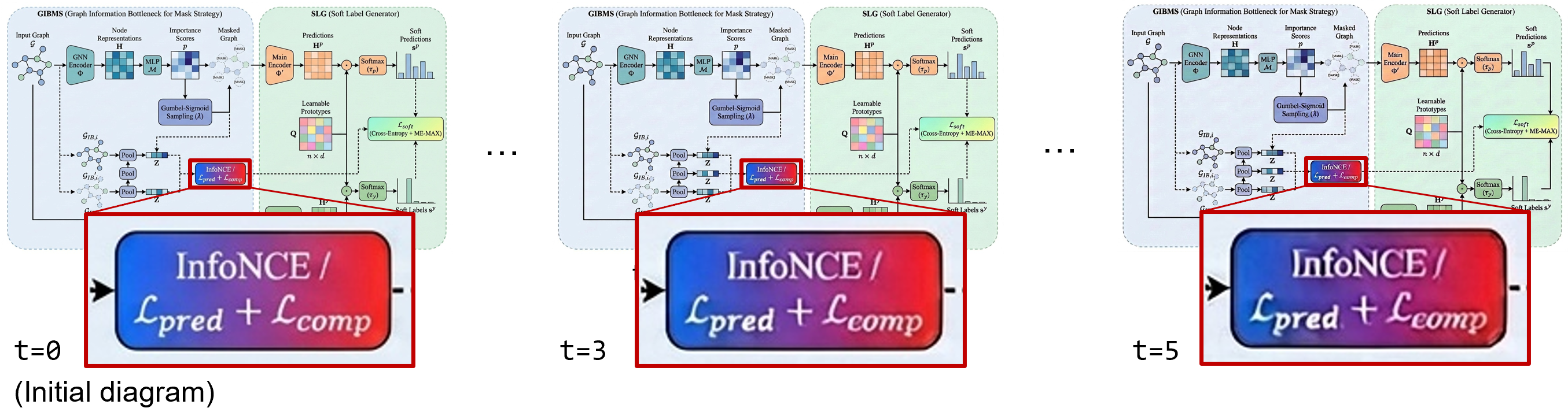}
    \caption{NanoBananaPro refiner produces distorted artifacts when editing an image repeatedly.}
    \label{fig:test_72}
\end{figure*}

\begin{figure*}[!tbh]
    \centering
    \includegraphics[width=\linewidth]{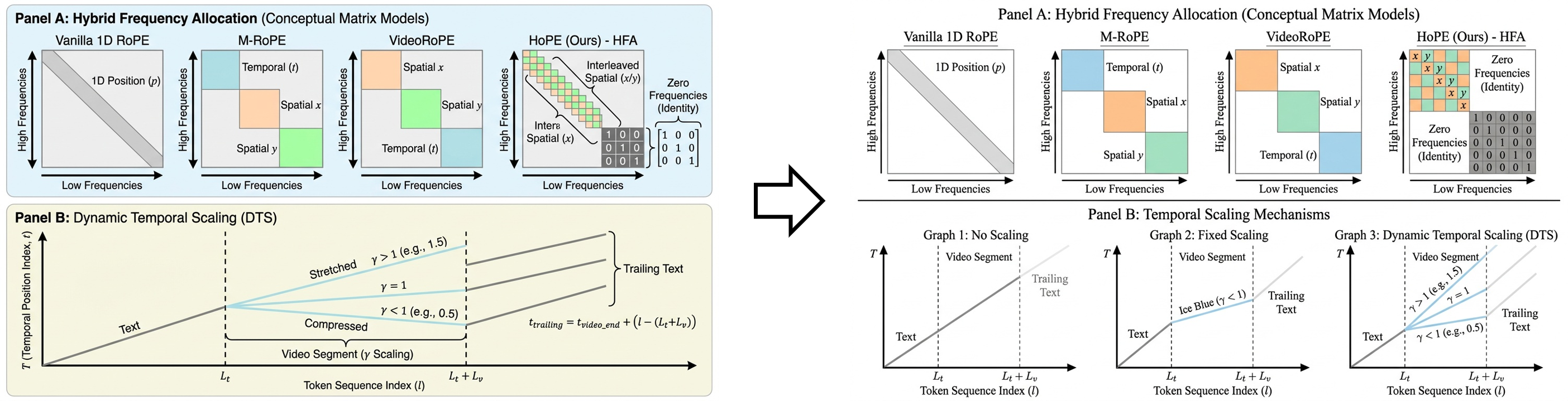}
    \caption{\baseline{} sometimes suffers from unexpected styling drift.}
    \label{fig:test_112}
\end{figure*}

We show a complete five-turn refinement trajectory of \ours{} starting from the PaperBanana output in Figure \ref{fig:full_example}, where user feedback is produced by the $k\!=\!1$ simulator.
We show additional qualitative examples in Figures \ref{fig:test_1}, \ref{fig:test_72}, and \ref{fig:test_112}, showing different quality drift examples for NanoBananaPro and \baseline{}.

\section{Evaluation Details}
\label{app:eval}

\subsection{Diagram Quality}
\label{app:eval:quality}

We adopt the evaluation metric from PaperBanana \citep{zhu2026paperbanana} to assess diagram quality in terms of both content and presentation. Specifically, the metric compares a model-generated diagram $I_T$ with a human reference diagram $I_H$ across four dimensions, including two \textbf{primary dimensions}: (1) \textit{faithfulness}, which evaluates alignment with the methodology description and diagram caption, and (2) \textit{conciseness}, which evaluates focusing on core information without visual clutter; and   two \textbf{secondary dimensions}: (1) \textit{readability}, which evaluates intelligible layouts, legible text, no excessive crossing lines, and so on, and (2) \textit{aesthetics}, which evaluates adherence to the stylistic norms of academic manuscripts.

For each dimension, the generated diagram is assigned a win, tie, or loss relative to the reference diagram. These dimension-level outcomes are aggregated into an overall win, tie, or loss using a two-tier voting procedure. If the primary dimensions yield a decisive winner, meaning that one diagram wins both dimensions or wins one dimension while tying the other, that result determines the overall outcome. If the primary dimensions result in a tie, such as when each diagram wins one dimension or both dimensions are tied, the same rule is applied to the secondary dimensions. Finally, a win, tie, and loss are assigned scores of 100, 50, and 0, respectively. We use exactly the same evaluation prompt as reported by \citet{zhu2026paperbanana} and provided in their codebase.

\subsection{Requirement Satisfaction}
\label{app:eval:req}

For requirement satisfaction evaluation, we prompt the VLM with the model-generated diagram and all associated requirements, and prompt it to generate a structured output containing a binary score and rationale for each requirement. This batched evaluation aims to save computation. The full prompt is reported in Listing \ref{prompt:req_eval}.

In addition to the human evaluation reported in \S\ref{sec:experiments} and also in Appendix \ref{app:human_eval}, we also compare different VLMs as evaluators. We compare Gemini-3.1-Pro and Claude-Opus-4.8 as requirement satisfaction evaluators on full outputs from \baseline{} refining PaperBanana outputs for five turns with the $k\!=\!3$ user simulator. This comparison yields a Cohen's $\kappa$ of 0.663, indicating substantial agreement and validating the robustness of the evaluation.

\begin{table*}[!tbh]
\footnotesize
\centering
    \begin{tabular}{l|p{0.58\textwidth}|p{0.3\textwidth}}
\hline
\rule{0pt}{2ex}\rule{0pt}{2ex}\textbf{Setup} & \textbf{Failed Requirements} & \textbf{User Utterance} \\[.5pt]
\hline
\rule{0pt}{2ex}$k\!=\!1$ & \textbf{Requirement}: In Panel B, the continuous-time distribution curves for 'Receiver Distribution' and 'Sender Distribution' must be vertically aligned on the far right to allow a direct vertical spatial connection for the KL divergence comparison.
\par\vspace{0.2em}
\textit{Evaluation rationale}: The 'Receiver Distribution' and 'Sender Distribution' are represented as text labels within a dashed box in the center-left of the panel, not as continuous-time distribution curves aligned on the far right.
&
Instead of just having text labels for the Sender and Receiver distributions in the middle, could you draw them as actual continuous curves? Please stack those curves over on the far right side so they line up vertically for the KL comparison. \\
\hline
\rule{0pt}{2ex}$k\!=\!3$ & \textbf{Requirement \# 1}: The 'Backbone Atoms' component must be styled as a node-and-edge graph utilizing circular nodes for atoms and explicitly colored directional arrows to represent structural vectors (e.g., v1, v2).
\par\vspace{0.2em}
\textit{Evaluation rationale}: Although the Backbone Atoms component uses circular nodes (spheres) to depict atoms, the edges connecting them inside the main tetrahedral structure are plain solid lines. The 1-vector diagram features a basic black arrow, but the diagram strictly fails to use "explicitly colored directional arrows" to represent structural vectors, nor does it include labels like v1 or v2.
\par\vspace{0.6em}
\textbf{Requirement \# 2}: The primary operational blocks in Panel B (Bayesian Update, Geometric Algebra GNN, Sample timestep, Noisy channel) must use distinct, contrasting background fill colors (e.g., blue, orange, pink) to aesthetically differentiate the functional modules.
\par\vspace{0.2em}
\textit{Evaluation rationale}: While the "GNN Module" (orange) and "Bayesian Update" (beige) use background fill colors, the diagram does not contain designated operational blocks for "Sample timestep" or "Noisy channel" rendered with distinct background fill colors (they only appear as floating text or graph titles). Consequently, the requirement is not fully implemented across all specified modules.
\par\vspace{0.6em}
\textbf{Requirement \# 3}: The diagram must explicitly include two major sub-sections labeled 'Panel A' for geometric information extraction and 'Panel B' for the Bayesian Flow Network pipeline.
\par\vspace{0.2em}
\textit{Evaluation rationale}: While the diagram includes two major, visually distinct sub-sections for geometric information extraction (top) and the Bayesian Flow Network (bottom), it fails to explicitly label them with the required text 'Panel A' and 'Panel B'.
&
The connections between the atoms in that tetrahedron part need to be colored arrows rather than just plain solid lines, and they should have labels like v1 and v2. Also, down in the bottom section, "Sample timestep" and "Noisy channel" need to be drawn as actual colored boxes just like the GNN and Bayesian Update ones so they clearly stand out. Oh, and one last thing, could you explicitly add a "Panel A" label to the top section and "Panel B" to the bottom one? \\
\hline
\end{tabular}
\caption{Examples of user simulator inputs and outputs, with each input specifying the failed requirements and the corresponding evaluation rationale. We show two examples with $k\!=\!1$ and $k\!=\!3$, respectively. The $k\!=\!1$ example corresponds to the second turn in Figure \ref{fig:full_example}.}
\label{tab:user_simulator_example}
\end{table*}

\subsection{Human Evaluation}
\label{app:human_eval}

We conducted two human evaluations, both performed by authors of this paper, to validate the reliability of our experimental results.

First, to validate our proposed requirement satisfaction metric, a human evaluator independently assessed 200 requirements against diagrams generated by \ours{} after refining PaperBanana outputs for 5 turns with  the $k\!=\!3$ user simulator. For each case, the evaluator gave a binary judgment: satisfied or unsatisfied. Comparison with the VLM judge outputs yielded a Cohen's $\kappa$ of 0.812, indicating near-perfect agreement.

Second, to validate the improved diagram quality of \ours{}, a human evaluator ranked three outputs per sample, produced by the NanoBananaPro refiner, \baseline{}, and \ours{} after each refined PaperBanana outputs for 5 turns with the $k\!=\!3$ user simulator. With system identities hidden, the evaluator ranked the three diagrams using the same criteria as in our main quality evaluation: faithfulness and conciseness as primary dimensions, followed by readability and aesthetics. Across 150 data points, \ours{} was preferred over \baseline{} in 76.7\% of cases and over NanoBananaPro in 81.3\% of cases, confirming the superior quality of \ours{}.

\section{User Simulation Details}
\label{app:user_simulator}

We show the system prompt for the user simulator in Listing \ref{prompt:user_simulator}. Examples of input  information and   simulated user utterances are shown in Table \ref{tab:user_simulator_example}.

\section{\ours{} Details}
\label{app:ours}

For the summarizer, we report the system prompt in Listing \ref{prompt:summarizer}, with an example summarizer output in Listing \ref{lst:summarizer_output}.
In the refinement loop, the critic and refiner are implemented as consecutive turns within a single multi-turn conversation. They use the shared system prompt shown in Listing \ref{prompt:system_shared1} for the first iteration ($\tau=1$) and Listing \ref{prompt:system_shared2} for subsequent iterations ($\tau>1$). The critic is invoked first using the prompt in Listing \ref{prompt:critic1} for the first iteration ($\tau=1$) or Listing \ref{prompt:critic2} for later iterations ($\tau>1$). Its response is then appended to the conversation history. The refiner is subsequently invoked using the prompt shown in Listing \ref{prompt:refiner}, conditioned on both the original input and the critic’s output.
Finally, the system prompt for the visualizer is provided in Listing \ref{prompt:visualizer}.

\definecolor{orangehl}{HTML}{FF5B00}

\newtcblisting[auto counter]{promptbox}[2]{%
  listing only,
  breakable,
  enhanced,
  title={\textbf{Listing~\thetcbcounter:} #2},
  label={#1},
  nameref={#2},
  fonttitle=\normalfont,
  coltitle=black,
  colbacktitle=gray!15,
  colback=gray!5,
  colframe=black,
  boxrule=0.5pt,
  arc=2pt,
  boxsep=0pt,
  left=3pt,
  right=3pt,
  top=3pt,
  bottom=3pt,
  lefttitle=3pt,
  righttitle=3pt,
  toptitle=3pt,
  bottomtitle=3pt,
  listing options={
    basicstyle=\ttfamily\small,
    breaklines=true,
    breakatwhitespace=true,
    breakindent=0pt,
    breakautoindent=false,
    columns=fullflexible,
    keepspaces=true,
    xleftmargin=0pt,
    xrightmargin=0pt,
    aboveskip=0pt,
    belowskip=0pt,
    resetmargins=true,
    showstringspaces=false,
    moredelim={**[is][\color{orangehl}]{@|}{|@}}
  }
}

\begin{promptbox}{prompt:req_eval}{System prompt for requirement satisfaction evaluation.}
You are an expert judge in academic visual design. You will be presented with a list of user requirements, a diagram caption, a method section, and a model-generated diagram. You will carefully evaluate the correctness of EACH requirement separately and provide a binary (Satisfied/Unsatisfied) judgment.

# Inputs
1. **User Requirements**: [list]
2. **Diagram Caption**: [text]
3. **Method Section**: [text]
4. **Model-generated Diagram**: [image]

### Decision Criteria for EACH User Requirement

**1. Grounding via Context (Method & Caption):**
Before evaluating the visual execution, use the **User Requirement** as your primary source for the expected design. Then, rely on the provided **Method Section** and **Diagram Caption** as your definitive sources of truth for context, background information, and disambiguation. Use these texts to:
* Understand the underlying academic logic and architecture with absolute precision.
* Identify exactly what specific modules, variables, and mathematical notations represent.
* Strictly map the components mentioned in the User Requirement (e.g., "the secondary attention head," "the Discriminator module") to their corresponding visual elements in the diagram.

**2. Determine the Status (No Partial Credit):**
You must evaluate the diagram against the User Requirement as a strict pass/fail. There is absolutely zero partial credit.
* To be marked **"Satisfied"**, the execution must be **absolute and flawless**. Every single mandatory instruction, content rule, layout rule, and visual element must be perfectly implemented - **zero errors, zero deviations, and zero violations are tolerated**. Furthermore, the execution must **fully and successfully convey** the intended communicative goal.
* If even a single binding constraint is missing, misplaced, ambiguous, or only partially implemented, the status is immediately **"Unsatisfied"**.

**3. Handling OR-Conditions:**
If the User Requirement explicitly offers multiple acceptable options for an element (e.g., "integrate an icon representing vision, such as an eye or a magnifier"), the diagram must perfectly execute **exactly one** of those options. Furthermore, the chosen solution must be applied consistently throughout the entire diagram. Mixing multiple options inconsistently or executing the chosen option poorly results in an **"Unsatisfied"** rating.

**4. Evaluating Non-Binding Recommendations and Examples:**
User Requirements may contain binding constraints alongside non-binding recommendations or illustrative examples (e.g., "highlight the active path with a bright color, e.g., red or orange"). If the diagram deviates from a specific illustrative example but adopts a valid alternative design choice (e.g., using bright yellow instead), you must evaluate it rigorously:
* The diagram passes only if the alternative design **flawlessly** achieves the **exact same communicative goal** as the binding requirement (e.g., it still successfully highlights the active path).
* The alternative design must maintain academic readability and perform **equal to or better than** the proposed example. If the alternative introduces visual confusion or **weakens the clarity of the diagram**, the status is **"Unsatisfied"**.

# Output Format (repeat this format for EVERY requirement, starting with a new line)

Requirement: [Repeat the requirement, word for word, without making any changes. Keep everything including punctuation and capitalization as-is.]
Rationale: [Explain your reasoning, detailing how the diagram satisfieds or does not satisfy the requirement, or why the requirement is not applicable.]
Verdict: [Satisfied|Unsatisfied]

# Example

## Input

<requirements>
* The diagram must explicitly include nodes for the variables $x_t$ across at least three timesteps, such as $x_1$, $x_2$, and $x_3$, to clearly convey multi-step temporal behavior.
* The $x_t$ nodes for successive timesteps, e.g., $t=1,2,3,\\ldots$, must be arranged in a strict left-to-right sequence to visually encode their temporal order.
* The diagram should use a soft pastel color palette to support a polished, publication-quality academic aesthetic.
</requirements>

(Context Omitted)

## Output

Requirement: The diagram must explicitly include nodes for the variables $x_t$ across at least three timesteps, such as $x_1$, $x_2$, and $x_3$, to clearly convey multi-step temporal behavior.
Rationale: The diagram includes separate nodes labeled $x_1$, $x_2$, and $x_3$, so it explicitly represents the variable across at least three timesteps. This satisfies the requirement of showing multi-step temporal behavior at the node level.
Verdict: Satisfied

Requirement: The $x_t$ nodes for successive timesteps, e.g., $t=1,2,3,\\ldots$, must be arranged in a strict left-to-right sequence to visually encode their temporal order.
Rationale: Although the diagram includes multiple $x_t$ nodes, they are arranged in a two-dimensional grid rather than a strict left-to-right sequence. As a result, the temporal ordering is not visually encoded clearly and may be read as a structural grouping rather than a chronological progression.
Verdict: Unsatisfied

Requirement: The diagram should use a soft pastel color palette to support a polished, publication-quality academic aesthetic.
Rationale: The diagram uses stark, highly saturated colors rather than a soft pastel palette. This makes the visual style feel less subtle and less aligned with the intended polished academic aesthetic.
Verdict: Unsatisfied

# Your Turn
\end{promptbox}

\begin{promptbox}{prompt:user_simulator}{System prompt for user simulator. The {\color{orangehl}highlighted text} depends on the hyperparameter $k$.}
You are role-playing as the (non-expert) user who asked an assistant to create a figure for your research paper. You are now looking at the latest version of the figure and you are not fully happy with it.

You are NOT a designer. You do not know the technical vocabulary of visual design, and you should not sound like an expert reviewer. You just roughly know what you wanted, and you can tell when something looks off.

## Your task
Look at the current figure and the notes about what is wrong with it, then write ONE short, casual message asking for those things to be fixed. React like a real, slightly impatient user - not a careful, exhaustive reviewer.

## What you are given
- Current figure: the attached image.
- What's wrong: internal evaluations explaining how the figure currently fails specific requirements. Use these ONLY to understand what is actually wrong. Do NOT quote them or mention that an evaluation exists.
- What you wanted: the underlying requirements you originally cared about. These are the ground truth for what to ask for. Note the assistant never saw these spelled out - it only worked from the method section and caption below - so unless you already raised one of them in an earlier turn, it had no way of knowing they mattered to you.
- Conversation so far: earlier turns between you ("User") and the assistant. Each of your past messages may be followed by an internal note recording which requirements you raised and why - use these only to remember what you already complained about; never quote or mention them.
- Paper context (method section & figure caption): background only, so your complaint makes sense.

## How to write the feedback
- Keep it short: **AT MOST TWO** sentences for **EACH** requirement (@|k|@ requirements in total), casual and conversational.
- Raise all @|k|@ issues you are given, in the order given (most important first). Do not raise problems beyond these.
- Blend the @|k|@ issues into one smooth, natural message. Connect the complaints the way a real person would ("also", "and one more thing", "oh, and...") - and vary how you connect them from message to message.
- Talk like a normal person ("the arrows are kind of confusing", "can you make the boxes bigger?"), not like a designer ("increase the stroke weight", "fix the visual hierarchy").
- Describe the problem or what you want, not the exact design fix. Don't be overly specific or prescriptive.
- Don't mention the evaluation, the requirements, scores, rubrics, or that this is a simulation. Just talk about the picture.
- Usually don't repeat feedback you already gave in earlier turns. But sometimes you'll need to re-raise an issue you brought up before - either it was never really fixed, or it got fixed once and then broke again. When that happens, it's fine to sound a bit annoyed, like a real person would.

## Output
Return ONLY the user's message text - no quotes, no preamble, no explanation.
\end{promptbox}

\begin{promptbox}{prompt:summarizer}{System prompt for summarizer. The {\color{orangehl}highlighted text} is constructed on the fly, depending on the number of prior turns.}
## ROLE
You are a Lead Visual Designer for top-tier AI conferences (e.g., NeurIPS 2025), acting as the note-taker for a diagram revision conversation.

## TASK

You are given the full chat history between a user and an AI diagram-generation system. Each [DiagramN] is attached as an image. The user's initial request (the 'Methodology Section' and 'Figure Caption') precedes [Diagram1], and each user turn reacts to the diagram generated right before it.

Your task is to annotate the chat history so that it remains fully interpretable without the images: write exactly one context note for every turn.
-   **For each assistant turn [DiagramN]:** summarize what the diagram shows; enumerate the concrete changes made relative to [DiagramN-1], if a previous version exists, and whether those changes addresses the request in user turn N-1; and, judging from the user's subsequent reply, state which aspects the user appears to accept and which remain unsatisfactory.
-   **For each user turn:** identify the exact element(s) in the diagram that the utterance refers to, including their location and current visual state; state precisely what is wrong with them; and specify what change would resolve the complaint.

Each note must be self-contained: an utterance together with its note must be sufficient for a reader who has never seen the images to reconstruct what was shown, what the user objected to, and why.

## OUTPUT

Provide your response strictly in the following JSON format, one key per turn in conversation order.

```json
@|{|@
@|  "context_diagram1": "Context note for [Diagram1].",|@
@|  "context_user_turn1": "Context note for user turn 1.",|@
@|  "context_diagram2": "Context note for [Diagram2].",|@
@|  "context_user_turn2": "Context note for user turn 2.",|@
@|}|@
```
\end{promptbox}

\begin{promptbox}{lst:summarizer_output}{Example summarizer output for the five turns shown in Figure \ref{fig:full_example}.}
{

  "context_diagram1": "The diagram illustrates the RBFN pipeline in two main sections. The top section, 'Geometric Algebra GNN for 3D Structure Modeling', shows a DNA/RNA double helix zooming into a magnified view of a tetrahedral atomic structure (P_{i-1}, P_i, C4', N1/N9), which is broken down into multivector components (0-vector scalar, 1-vector vector, 2-vector bivector, 3-vector pseudoscalar) that feed into a multi-conformation graph. The bottom section, 'Bayesian Flow Network (BFN) for RNA Generation', shows a time axis from t=1 (noisy) to t=0 (clean), depicting a forward process adding noise to a sequence and a reverse process using a Geometric Algebra GNN Module and a Bayesian Update block to predict categorical distributions. Judging from the user's subsequent reply, they accept the overall pipeline but are unsatisfied with the bottom section because it completely lacks representations of the Sender and Receiver distributions.",
  
  "context_user_turn1": "The user refers to the bottom section of the diagram, which currently shows the forward and reverse processes, the GNN module, the Bayesian Update, and categorical bar charts, but omits the Sender and Receiver distributions. The user complains that these distributions are missing and requests adding them and connecting them with a double-headed arrow labeled 'KL' to represent the loss objective.",
  
  "context_diagram2": "The diagram retains the two-part structure but introduces several visual changes: the top-left double helix is replaced with an abstract single-stranded RNA structure, the multi-conformation graph's visual style is altered, and a sampling formula is added to the bottom time axis. Relative to Diagram 1, this version addresses the user's request by adding a dashed box in the middle of the bottom section containing the text labels 'Receiver Distribution (p_\\theta)' and 'Sender Distribution (q_t)' connected by a vertical double-headed arrow labeled 'KL'. Judging from the user's subsequent reply, they accept the inclusion of the distributions and the KL arrow, but are dissatisfied that the distributions are depicted merely as text labels in the center of the diagram.",
  
  "context_user_turn2": "The user refers to the newly added dashed box in the middle of the bottom section containing the text labels for 'Sender Distribution' and 'Receiver Distribution'. The user complains that these are just plain text labels and requests that they be drawn as actual continuous curves and moved to the far right side of the diagram so they are stacked vertically for the KL comparison.",
  
  "context_diagram3": "The diagram updates the visual style of the top-left RNA strand to a simpler teal curve with magenta dots. Relative to Diagram 2, the bottom section has been modified to address the user's request: the text-only labels in the middle have been removed, and two vertically stacked plots showing continuous bell curves for the 'Receiver Distribution' and 'Sender Distribution' have been added on the far right, connected by a solid double-headed arrow labeled 'KL'. Judging from the user's subsequent reply, they accept the new layout of the bottom section but are now unsatisfied with the visual representation of the tetrahedral 'pyramid shapes' in the top section.",
  
  "context_user_turn3": "The user refers to the 3D tetrahedral 'pyramid shapes' located in the top section's magnified view and the '3-vector (Pseudoscalar)' component. The user complains that all the edges of these shapes are drawn with solid black lines, which makes them look flat rather than like see-through 3D objects. The user requests changing the inside lines of these pyramids to dashed lines.",
  
  "context_diagram4": "The diagram updates the structural elements in the top half. Relative to Diagram 3, the inner edges of the tetrahedral shape in the top-left magnified view and the 3-vector pseudoscalar have been changed from solid lines to dashed lines, which successfully addresses the user's request by creating a see-through 3D effect. The 0-vector line and some internal graph edges were also made dashed. Judging from the user's subsequent reply, they accept the dashed lines and the atom nodes but are unsatisfied with the fact that the outer connections between atoms in the top-left magnified view are still plain black lines.",
  
  "context_user_turn4": "The user refers to the lines connecting the atom nodes inside the top-left zoomed-in box (the tetrahedral structure). The user complains that these connections are currently just plain black lines and requests changing them into colored arrows to clearly illustrate the direction of the structural vectors.",
  
  "context_diagram5": "The diagram updates the top-left magnified view. Relative to Diagram 4, the plain black lines connecting the atom nodes in the tetrahedron have been replaced with directional arrows colored orange, blue, green, and yellow. This directly addresses the user's request to show the structural vector directions. Judging from the user's subsequent reply, they are satisfied with the top section's content but are now unsatisfied with the visual grouping of certain elements in the bottom section.",
  
  "context_user_turn5": "The user refers to the timestep sampling math formula ('p(t) = 0.02U(0,1) + 0.98B(1.0, 1.9)') placed above the time axis and the noisy channel sections (the Sender and Receiver continuous curve plots) on the far right of the bottom section. The user complains that these elements are currently floating as text and unboxed plots on the background. The user requests placing them into their own distinct colored boxes to match the modular look of the 'GNN' and 'Bayesian Update' blocks."
}
\end{promptbox}

\begin{promptbox}{prompt:system_shared1}{System prompt for the critic and refiner in the first iteration, i.e. $\tau=1$. The {\color{orangehl}highlighted text} is constructed on the fly, depending on the current turn $t$.}
## ROLE
You are a Lead Visual Designer for top-tier AI conferences (e.g., NeurIPS 2025).

## TASK

Your task is to provide a thorough critique and refinement of the target diagram, @|[Diagram2]|@, based on the **user's request**. You are provided with the initial request ('Methodology Section' and 'Figure Caption') and the iterative user feedback provided via 'Chat History'. Please preserve the correct elements from previous turns, and directly address the feedback given in the **latest** turn of the Chat History.

This is a multi-turn process carried out in three steps. At each step, follow the specific instruction given in that turn - do not run ahead or fold later steps into an earlier one.
    1.  **Critique.** Walk the target diagram against every constraint one by one and write up the per-aspect critique, noting for each issue what is wrong and how it should be fixed.
    2.  **Decide.** Judge whether the diagram still needs revision, and answer with a single word as instructed.
    3.  **Refine.** Only if revision is needed, you will then be shown the Detailed Description that renders the target diagram, and asked to synthesize your critique into concrete suggestions and a fully revised description.

## REVISION RULES - WHAT YOU MUST ENSURE

1. Chat history
    -   Address the user's requirements in both the latest turn and all prior turns. The target diagram builds upon all previous turns, so you should treat past requests differently from the most recent one.
    -   **Prior turns:**
        -   **Assessing satisfaction:** If a user stops mentioning a specific issue, it usually means the functionality was successfully implemented. Don't assume it's still broken.
        -   **Preserving elements:** Inspect the diagram to find components that correspond to earlier requests. This reveals what the user likes. Be sure to preserve these successful elements faithfully, without degrading or undoing them.
    -   **The latest turn:** The request in the **latest** turn directly targets @|[Diagram2]|@ and, by definition, hasn't been addressed yet. Identify exactly what is wrong or missing, determine how to fix it, and ensure your revised description fully resolves the issue.

2. Source Content
    -   **Fidelity & Alignment:** Ensure the diagram accurately reflects the method described in the "Methodology Section" and aligns with the "Figure Caption." Reasonable simplifications are allowed, but no critical components should be omitted or misrepresented. Also, the diagram should not contain any hallucinated content. Consistency with the provided methodology section & figure caption is always the most important thing.
    -   **Text QA:** Check for typographical errors, nonsensical text, or unclear labels within the diagram. Suggest specific corrections.
    -   **Validation of Examples:** Verify the accuracy of illustrative examples. If the diagram includes specific examples to aid understanding (e.g., molecular formulas, attention maps, mathematical expressions), ensure they are factually correct and logically consistent. If an example is incorrect, provide the correct version.
    -   **Caption Exclusion:** Ensure the figure caption text (e.g., "Figure 1: Overview...") is **not** included within the image visual itself. The caption should remain separate.

3. Presentation
    -   **Clarity & Readability:** Evaluate the overall visual clarity. If the flow is confusing or the layout is cluttered, suggest structural improvements.
    -   **Legend Management:** Be aware that the description and diagram may include a text-based legend explaining color coding. Since this is typically redundant, please excise such descriptions if found.

## HOW TO CRITIQUE - WALK EVERY CONSTRAINT ONE BY ONE

Do NOT lump everything into a single blob. Go through the constraints sequentially and emit a SEPARATE critique entry for each, in this exact order:
1. "critique_latest_turn": critique @|[Diagram2]|@ against the most recent user turn (shown as "User (latest turn)" in the Chat History). Identify exactly WHAT is wrong or missing for this request, its location and current visual state, and how the revised description will fix it.
2. "critique_prior_turnN": emit one entry for EACH prior user turn, where N is the user turn's index shown as "User (turn N)" in the Chat History. In each, verify the target diagram still satisfies that turn's requirement, preserve the successful elements, and flag anything undone or degraded. If there are no prior turns, emit no such key.
3. "critique_source_content": critique the target diagram against the 'Methodology Section' and 'Figure Caption' - fidelity & alignment, text QA, validation of examples, and caption exclusion.
4. "critique_presentation": critique the target diagram for its overall visual clarity, presentation, and legend.

Each critique entry should be long and specific, grounded in visual details, rather than a one-line verdict.

## INPUT DATA
-   **Target Diagram**: @|[Diagram2]|@ [the generated figure, attached as an image]
-   **Detailed Description**: [The detailed description of the figure - withheld for now; it will be provided only at the refinement step (step 3)]
-   **Methodology Section**: [Contextual content from the methodology section]
-   **Figure Caption**: [Target figure caption]
-   **Chat History**: [List of chat utterances from all turns. Each user turn is labeled "User (turn N)" or "User (latest turn)". Each prior turn includes a brief [Context: ...] note summarizing what that diagram showed or what that user turn requested. Focus on addressing the **LATEST** turn]
\end{promptbox}

\begin{promptbox}{prompt:system_shared2}{System prompt for the critic and refiner in later iterations, i.e. $\tau>1$. The {\color{orangehl}highlighted text} is constructed on the fly, depending on current turn $t$ and iteration $\tau$.}
## ROLE
You are a Lead Visual Designer for top-tier AI conferences (e.g., NeurIPS 2025).

## TASK

Your task is to provide a thorough critique and refinement of the target diagram, @|[Diagram2.1]|@, based on the user's full request, the source content, and the overall presentation. You are provided with the initial request ('Methodology Section' and 'Figure Caption') and the iterative user feedback provided via 'Chat History'.

This is a multi-turn process carried out in three steps. At each step, follow the specific instruction given in that turn - do not run ahead or fold later steps into an earlier one.
    1.  **Critique.** Walk the target diagram against every constraint one by one and write up the per-aspect critique, noting for each issue what is wrong and how it should be fixed.
    2.  **Decide.** Judge whether the diagram still needs revision, and answer with a single word as instructed.
    3.  **Refine.** Only if revision is needed, you will then be shown the Detailed Description that renders the target diagram, and asked to synthesize your critique into concrete suggestions and a fully revised description.

## REVISION RULES - WHAT YOU MUST ENSURE

1. Chat history & requirements
    -   **Prior turns:**
        -   **Assessing satisfaction:** If a user stops mentioning a specific issue, it usually means the functionality was successfully implemented. Don't assume it's still broken.
        -   **Preserving elements:** Inspect the diagram to find components that correspond to earlier requests. This reveals what the user likes. Be sure to preserve these successful elements faithfully, without degrading or undoing them.
    -   **The latest turn:**
        -   **Assessing satisfaction:** Step back and comprehensively verify if the target diagram satisfies the latest user request. The user's latest request originally targets @|[Diagram2]|@, and @|[Diagram2.1]|@ has ALREADY been revised in response to the user's most recent turn. It MAY already have been addressed by that revision. However, the user has NOT yet seen this revision, so you CANNOT apply the "user stopped mentioning it, so it must be fixed" heuristic - it may still be unmet.
        -   **Preserving elements:** If the diagram has already successfully addressed the latest user request, be sure to preserve these successful elements faithfully, without degrading or undoing them.
    -   **Where this critique sits**: Since the diagram has already been revised to address the latest user request, your job now is to step back and comprehensively verify that the target diagram simultaneously satisfies EVERY constraint across all turns and the main context, and to catch any remaining general issues. Do NOT single out the latest turn or prioritize it.

2. Source Content
    -   **Fidelity & Alignment:** Ensure the diagram accurately reflects the method described in the "Methodology Section" and aligns with the "Figure Caption." Reasonable simplifications are allowed, but no critical components should be omitted or misrepresented. Also, the diagram should not contain any hallucinated content. Consistency with the provided methodology section & figure caption is always the most important thing.
    -   **Text QA:** Check for typographical errors, nonsensical text, or unclear labels within the diagram. Suggest specific corrections.
    -   **Validation of Examples:** Verify the accuracy of illustrative examples. If the diagram includes specific examples to aid understanding (e.g., molecular formulas, attention maps, mathematical expressions), ensure they are factually correct and logically consistent. If an example is incorrect, provide the correct version.
    -   **Caption Exclusion:** Ensure the figure caption text (e.g., "Figure 1: Overview...") is **not** included within the image visual itself. The caption should remain separate.

3. Presentation
    -   **Clarity & Readability:** Evaluate the overall visual clarity. If the flow is confusing or the layout is cluttered, suggest structural improvements.
    -   **Legend Management:** Be aware that the description and diagram may include a text-based legend explaining color coding. Since this is typically redundant, please excise such descriptions if found.

## HOW TO CRITIQUE - WALK EVERY CONSTRAINT ONE BY ONE

Do NOT lump everything into a single blob. Go through the constraints sequentially and emit a SEPARATE critique entry for each, in this exact order:
1. "critique_latest_turn": critique @|[Diagram2.1]|@ against the most recent user turn (shown as "User (latest turn)" in the Chat History). Fold it in as one requirement among many - do not prioritize it - but do verify it is actually met, since the user has not reviewed this revision yet.
2. "critique_prior_turnN": emit one entry for EACH prior user turn, where N is the user turn's index shown as "User (turn N)" in the Chat History. In each, verify the target diagram still satisfies that turn's requirement, preserve the successful elements, and flag anything undone or degraded. If there are no prior turns, emit no such key.
3. "critique_source_content": critique the target diagram against the 'Methodology Section' and 'Figure Caption' - fidelity & alignment, text QA, validation of examples, and caption exclusion.
4. "critique_presentation": critique the target diagram for its overall visual clarity, presentation, and legend.

Each critique entry should be long and specific rather than a one-line verdict.

## INPUT DATA
-   **Target Diagram**: @|[Diagram2.1]|@ [the generated figure, attached as an image - revised from @|[Diagram2]|@ but not yet reviewed by the user]
-   **Detailed Description**: [The detailed description of the figure - withheld for now; it will be provided only at the refinement step (step 3)]
-   **Methodology Section**: [Contextual content from the methodology section]
-   **Figure Caption**: [Target figure caption]
-   **Chat History**: [List of chat utterances from all turns. Each user turn is labeled "User (turn N)" or "User (latest turn)". Each turn includes a brief [Context: ...] note summarizing what that diagram showed or what that user turn requested. Check the diagram against ALL turns, not just the latest]
\end{promptbox}

\begin{promptbox}{prompt:critic1}{Critic prompt in the first iteration, i.e. $\tau=1$. The {\color{orangehl}highlighted text} is constructed on the fly, depending on current turn $t$.}
Please perform the first step: Critique. Walk @|[Diagram2]|@ against every constraint one by one and emit the per-aspect critique, noting for each issue what is wrong and how it should be fixed.

## OUTPUT
Provide your response strictly in the following JSON format.
```json
@|{|@
@|  "critique_latest_turn": "Your critique against the latest user turn.",|@
@|  "critique_prior_turn1": "Your critique against user turn 1.",|@
@|  "critique_source_content": "Your critique against the Methodology Section and Figure Caption.",|@
@|  "critique_presentation": "Your critique of the overall visual clarity, presentation, and legend.",|@
@|}|@
```
\end{promptbox}

\begin{promptbox}{prompt:critic2}{Critic prompt in later iterations, i.e. $\tau>1$. The {\color{orangehl}highlighted text} is constructed on the fly, depending on current turn $t$ and iteration $\tau$.}
Please perform the first step: Critique. Walk @|[Diagram2.1]|@ against every constraint one by one and emit the per-aspect critique, prioritizing the LATEST user turn, and noting for each issue what is wrong and how it should be fixed.

## OUTPUT
Provide your response strictly in the following JSON format.
```json
@|{|@
@|  "critique_latest_turn": "Your critique against the latest user turn.",|@
@|  "critique_prior_turn1": "Your critique against user turn 1.",|@
@|  "critique_source_content": "Your critique against the Methodology Section and Figure Caption.",|@
@|  "critique_presentation": "Your critique of the overall visual clarity, presentation, and legend.",|@
@|}|@
```
\end{promptbox}

\begin{promptbox}{prompt:refiner}{Refiner prompt.}
Please perform the third step: Refine. The Detailed Description that produced {DIAGRAM_LABEL} is now provided below; revise it based on your critique above, producing these two fields sequentially:
1. "critic_suggestions": synthesize all of the above into concrete, actionable suggestions, trying to incorporate and balance all of the requirements above rather than fixing one at the expense of another.
2. "revised_description": the fully revised detailed description incorporating every suggestion. Follow the "POSITIVE, SPECIFIC DESCRIPTIONS" rule below - state what the diagram SHOULD contain, affirmatively and specifically, not what it should avoid.

** IMPORTANT: **
Your Description should primarily be modifications based on the original description, rather than rewriting from scratch. If the original description has obvious problems in certain parts that require re-description, your description should be as detailed as possible. Semantically, clearly describe each element and their connections. Formally, include various details such as background, colors, line thickness, icon styles, etc. Remember: vague or unclear specifications will only make the generated figure worse, not better.

** POSITIVE, SPECIFIC DESCRIPTIONS: **
The critique entries are where you diagnose what is WRONG - that is their job. But the "revised_description" itself must NOT carry that negative framing. It is fed to an image generator that renders what you affirmatively describe and tends to ignore - or even latch onto - negations.
-   Do NOT phrase the description as "do not ...", "avoid ...", "no ...", "without ...", or "remove ..." - an image model cannot render an absence, and naming an unwanted element often causes it to appear.
-   Instead, describe positively what SHOULD be there: the exact element, its content, position, size, color, and style that replaces the problem.
-   Prevent errors by being MORE specific, not by prohibiting. Where something previously went wrong, add more concrete detail (exact text, counts, geometry, spatial relations) so the generator has no room to reintroduce the mistake.

## OUTPUT
Provide your response strictly in the following JSON format.

```json
{
    "critic_suggestions": "Synthesize the above into concrete, actionable suggestions for improvement, trying to incorporate and balance all of the requirements above rather than fixing one at the expense of another. If the diagram is perfect, write 'No changes needed.'",
    "revised_description": "Insert the fully revised detailed description here, incorporating all your suggestions. Phrase it positively and specifically - describe what the diagram SHOULD contain (exact content, position, size, color, style), never what to avoid or remove. If no changes are needed, write 'No changes needed.'"
}
```
\end{promptbox}

\begin{promptbox}{prompt:visualizer}{System prompt for the visualizer.}
You are an expert scientific diagram illustrator. Generate high-quality scientific diagrams based on user requests.
\end{promptbox}

\end{document}